\documentclass{article}

\usepackage{amsmath,amsfonts,bm}

\def\eqref#1{equation~\ref{#1}}

\def\1{\bm{1}}

\def\vc{{\bm{c}}}

\DeclareMathAlphabet{\mathsfit}{\encodingdefault}{\sfdefault}{m}{sl}
\SetMathAlphabet{\mathsfit}{bold}{\encodingdefault}{\sfdefault}{bx}{n}

\def\gD{{\mathcal{D}}}

\DeclareMathOperator*{\argmax}{arg\,max}

\usepackage{microtype}
\usepackage{graphicx}
\usepackage{amsmath,amsfonts,bm,bbm}
\usepackage{subcaption}
\usepackage{algorithm}
\usepackage[noend]{algpseudocode}
\usepackage{placeins}
\usepackage{pifont}
\usepackage[hyphens]{url}
\usepackage{enumitem}
\usepackage{makecell}
\usepackage{booktabs} 
\usepackage{arydshln}
\newcommand{\cmark}{\ding{51}}%
\newcommand{\xmark}{\ding{55}}%

\usepackage{hyperref}

\usepackage[accepted]{icml2020}

\icmltitlerunning{Improving Molecular Design by Stochastic Iterative Target Augmentation}

\newtheorem{proposition}{Proposition}
\begin{document}

\twocolumn[
\icmltitle{Improving Molecular Design by Stochastic Iterative Target Augmentation}



\icmlsetsymbol{equal}{*}

\begin{icmlauthorlist}
\icmlauthor{Kevin Yang}{berkeley}
\icmlauthor{Wengong Jin}{mit}
\icmlauthor{Kyle Swanson}{cambridge}
\icmlauthor{Regina Barzilay}{mit}
\icmlauthor{Tommi Jaakkola}{mit}
\end{icmlauthorlist}

\icmlaffiliation{berkeley}{UC Berkeley}
\icmlaffiliation{mit}{MIT}
\icmlaffiliation{cambridge}{University of Cambridge}

\icmlkeywords{data augmentation, generative models, self-training, chemistry, molecular optimization}

\icmlcorrespondingauthor{Kevin Yang}{yangk@berkeley.edu}

\vskip 0.3in
]



\printAffiliationsAndNotice{}  

\begin{abstract}
Generative models in molecular design tend to be richly parameterized, data-hungry neural models, as they must create complex structured objects as outputs. Estimating such models from data may be challenging due to the lack of sufficient training data. In this paper, we propose a surprisingly effective self-training approach for iteratively creating additional molecular targets. We first pre-train the generative model together with a simple property predictor. The property predictor is then used as a likelihood model for filtering candidate structures from the generative model. Additional targets are iteratively produced and used in the course of stochastic EM iterations to maximize the log-likelihood that the candidate structures are accepted. A simple rejection (re-weighting) sampler suffices to draw posterior samples since the generative model is already reasonable after pre-training. We demonstrate significant gains over strong baselines for both unconditional and conditional molecular design. In particular, our approach outperforms the previous state-of-the-art in conditional molecular design by over 10\% in absolute gain. Finally, we show that our approach is useful in other domains as well, such as program synthesis. 

\end{abstract}

\section{Introduction}

The goal of molecular generation is to create molecules with the desired property profile. This task is a key component of pharmaceutical drug discovery, and has received intense attention in recent years,
yielding a wide range of proposed architectures~\cite{you2018graph,olivecrona2017molecular,popova2018deep,jin2019multi}. A common feature of these architectures is reliance on a large number of parameters to generate molecules, which are represented as complex graph-structured objects. As a result, these models require copious amounts of training data, consisting of molecules with their target properties.  Collecting such property data is often slow and expensive due to the required empirical measurements. 

Our challenge is to achieve high-quality molecular generation in data-sparse regimes. While semi-supervised methods for representation learning have demonstrated significant benefits in natural language processing and computer vision~\cite{edunov2018understanding,lee2013pseudo}, they are relatively under-explored in chemistry. In this paper, we propose a simple and surprisingly effective self-training approach for iteratively creating additional molecular targets. This approach can be broadly applied to any generative architecture, without any modifications.

Our stochastic iterative target augmentation approach, shown in Figure \ref{fig:train_filter}, builds on the idea that it is easier to evaluate the properties of candidate molecules than to generate those molecules. Thus a learned property predictor can be used to effectively guide the generation process.
To realize this idea, our method starts by pre-training the generative model on a small supervised dataset along with the property predictor.
The property predictor then serves as a likelihood model for filtering candidate molecules from the generative model.
Candidate generations that pass this filtering become part of the training data for the next training epoch.
Theoretically, this procedure can be viewed as one iteration of stochastic EM, maximizing the log-likelihood that the candidate structures are accepted.
As the generative model already produces reasonable samples after pre-training, a simple rejection (re-weighting) sampler suffices to draw posterior samples.
For this reason, it is helpful to apply the filter at test time as well, or to use the approach transductively\footnote{Allowing the model to access test set inputs (but not targets) during training.} to further adapt the generation process to novel test cases. 
The approach is reminiscent of self-training or reranking approaches employed with some success for parsing~\citep{mcclosky2006effective,charniak2016parsing}. However, in our case, it is the candidate generator that is complex while the filter is relatively simple and remains fixed during the iterative process.


\begin{figure*}[t] 
    \centering
    \includegraphics[width=0.85\textwidth]{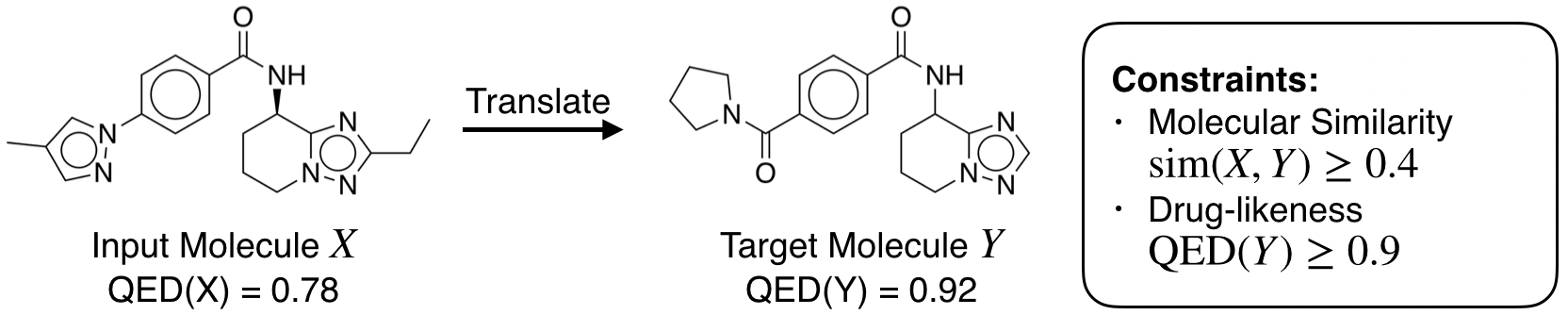}
    \caption{Illustration of conditional molecular design. Molecules can be modeled as graphs, with atoms as nodes and bonds as edges. Here, the task is to train a translation model to modify a given input molecule into a target molecule with higher drug-likeness (QED) score. The constraint has two components: the output $Y$ must be highly drug-like, and must be sufficiently similar to the input $X$.}
    \label{fig:molecule}
\end{figure*}

\begin{figure*}[t]
  \centering
  \includegraphics[width=\linewidth]{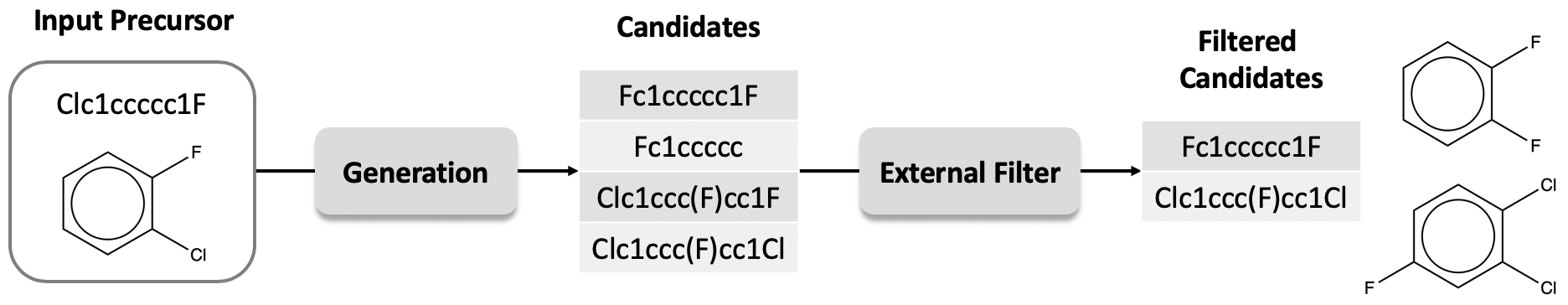}
  \caption{Illustration of data generation process for conditional molecular design. Given an input molecule, we first use our generative model to generate candidate modifications, and then select sufficiently similar molecules with high property score using our external filter. In the unconditional setting where the model takes no input, we simply sample outputs from the model and filter by property score.}
  \label{fig:train_filter}
\end{figure*}

We demonstrate that our target augmentation algorithm is effective and consistent across different generation tasks in its ability to improve molecular design performance. Our method is tested in two scenarios: molecular generative modeling (i.e., unconditional molecular design) and graph-to-graph translation, the corresponding conditional design problem of modifying an existing molecule to improve its properties. The latter is illustrated in Figure \ref{fig:molecule}.
We demonstrate significant gains over strong baselines for both settings. For instance, our approach outperforms the previous state-of-the-art~\citep{jin2019multi} in conditional molecular design by over 10\% in absolute gain on two tasks. 

Finally, our proposed method is not tied specifically to the molecular domain, and can generalize to any conditional or unconditional generation task with task-specific constraints. For example, in program synthesis, we show that our method outperforms a strong reinforcement learning baseline~\citep{bunel2018leveraging}. 



\section{Stochastic Iterative Target Augmentation}\label{method}

We present our method in the context of
conditional molecular design~\citep{jin2019multi,jin2018learning}, the task of transforming a given molecule $X$ into another compound $Y$ with improved chemical properties, while constraining $Y$ to remain similar to $X$ (Figure \ref{fig:molecule}). The corresponding unconditional task takes no input, seeking only to generate molecules with desired properties. 

As our method can be adapted to the unconditional setting by just dropping the input conditioning, we present our method in the conditional context. 
For a given input $X$, the model learns to generate an output $Y$ satisfying $\vc=1 | X, Y$ for some constraint $\vc$, represented as a binary random variable whose value is a function of $X$ and $Y$. (That is, $\vc$ corresponds to our filter.) For example, in conditional molecular generation, $\vc=1$ if $Y$ exceeds a specified property score threshold while being sufficiently similar to $X$. 
The proposed augmentation framework can be applied to any translation model $P$ trained on an existing dataset $\mathcal{D}=\{(X_i, Y_i)\}$, independent of the specific model architecture. 
As illustrated in Figure~\ref{fig:train_filter}, our method is an iterative procedure in which each iteration consists of the following two steps:
\begin{itemize}[leftmargin=*,topsep=0pt,itemsep=0pt] 
    \item \textbf{Augmentation Step}: Let $\gD$ be the original dataset and $\gD_t$ the training set at iteration $t$. To construct the next epoch's augmented training set $\gD_{t_1}$, we first initialize $\gD_{t+1} = \gD$. We then feed each input $X_i \in \gD$  into the translation model up to $C$ times to sample candidate translations $Y_i^1 \dots Y_i^{C}$.\footnote{One could initialize $\gD_{t+1} = \gD_{t}$ instead of $\gD_{t+1} = \gD$ and continuously expand the dataset, but the empirical effect is small (see Appendix \ref{more_molopt_exp}). Note our augmentation step can be trivially parallelized for speed.} We take the first $K$ distinct translations for each $X_i$ satisfying the constraint $\vc$ and add them to $\gD_{t+1}$. When we do not find $K$ distinct valid translations, we simply add copies of the original translation $Y_i$ to $\gD_{t+1}$ to preserve balance. In the unconditional setting, we instead just sample up to $C|\gD|$ outputs and accept up to $K|\gD|$ distinct new targets. 

    \item \textbf{Training Step}: We continue to train the model $P^{(t)}$ over the new training set $\gD_{t+1}$ for one epoch. 
\end{itemize}

The above training procedure is summarized in Algorithm~\ref{alg:algorithm}.
As the constraint $\vc$ is known a priori, we can construct an external property filter to remove generated outputs that violate $\vc$ during the augmentation step.
At test time, we also use this filter to screen predicted outputs. To propose the final translation of a given input $X$, we sample up to $L$ outputs from the model until we find one satisfying the constraint $\vc$. If all $L$ attempts fail for a particular input, we output the first of the failed attempts.

\begin{algorithm}[t] 
    \caption{Stochastic iterative target augmentation}
    \label{alg:algorithm}
    
    \textbf{Input:} Data $\gD=\{(X_1, Y_1), \dots, (X_n, Y_n)\}$, model $P^{(0)}$
    
    \begin{algorithmic}[1]
        \Procedure{AugmentDataset}{$\gD$, $P^{(t)}$}
            \State $\gD_{t+1} = \gD$ \Comment{Initialize augmented dataset}
            \For{$(X_i, Y_i)$ in $\gD$}
                \For{attempt in $1,\dots,C$}
                    \State Apply $P^{(t)}$ to $X_i$ to sample candidate $Y'$
                    \If{$\vc = 1 | X_i, Y'$ and $(X_i, Y')\notin \gD_{t+1}$}
                        \State Add $(X_i, Y')$ to $\gD_{t+1}$
                    \EndIf
                    \If{$K$ successful translations added} 
                        \State break from loop
                    \EndIf
                \EndFor
            \EndFor
            
            \State \Return{augmented dataset $\gD_{t+1}$}
        \EndProcedure
        \vspace{10pt}
        \Procedure{Train}{$\gD$}
            \For{epoch in $1, \dots, n_1$} \Comment{Regular training}
                \State Train model on $\gD$.
            \EndFor
            
            \For{epoch in $1, \dots, n_2$} \Comment{Augmentation}
                \State $\gD_{t+1} = $ \textsc{AugmentDataset($\gD$, $P^{(t)}$)}
                \State $P^{(t+1)} \leftarrow $ Train model $P^{(t)}$ on $\gD_{t+1}$.
            \EndFor
        \EndProcedure
    \end{algorithmic}
\end{algorithm}

Finally, as an additional improvement specific to the conditional setting, we observe that the augmentation step can be carried out for unlabeled inputs $X$ that have no corresponding $Y$. Thus we can further augment our training dataset in the transductive setting by including test set inputs during the augmentation step, or in the semi-supervised setting by simply including unlabeled inputs.


\section{Algorithm Motivation} \label{theory}

We provide here some theoretical motivation for our method in the conditional setting. 
Since molecules are discrete objects, we assume a discrete output space.


In the conditional context, the primary difficulty lies in generalizing to unseen inputs (precursors) at test time. Generating even a single successful $Y$ for a given $X$ is nontrivial. Therefore, we focus on maximizing the model's probability of generating successful translations.

We can characterize our method as a stochastic expectation-maximization (EM) algorithm~\citep{celeux1996stochastic}. 
As before, our external filter $\vc$ is a binary random variable whose value is a function of $X$ and $Y$, representing whether output $Y$ satisfies the desired constraint in relation to input $X$. We would like to generate $Y$ such that $Y \in B(X) \overset{def}{=} \{Y' : \vc = 1 | X, Y'\}$. If the initial translation model $P^{(0)}(Y|X)$ (after bootstrapping on the gold data, but before our augmentation) serves as a reasonable prior distribution over outputs $Y$ for any given input $X$, we could simply ``invert'' the filter and use
\begin{equation}\label{initial_posterior}
    P^{(*)}(Y|X) \propto P^{(0)}(Y|X) \cdot p(\vc=1 | X, Y)
\end{equation}
as the ideal translation model, noting that the probability $p(\vc=1 | X, Y)$ is either 0 or 1 since $\vc$ is a function of $X$ and $Y$. This posterior calculation is typically infeasible but can be approximated through sampling; even so, it relies heavily on the appropriateness of the prior $P^{(0)}(Y|X)$. Instead, we go a step further and iteratively optimize our parametrically defined prior translation model $P_{\theta}(Y|X)$. Note that the resulting prior can become much more concentrated around acceptable translations. 

We maximize the log-likelihood that candidate translations satisfy the constraints implicitly encoded in the filter: \begin{equation}
    \mathbb{E}_X \left[ \log P_\theta(\vc=1 \;|\; X) \right] \label{eq:objective}
\end{equation}

In many cases there are multiple viable outputs for any given input $X$. The training data may provide only one (or none) of them. Therefore, we treat the output structure $Y$ as a latent variable, and expand the inner term of Eq.(\ref{eq:objective}) as
\begin{align}
    \log \sum_Y P_\theta(Y | X)\cdot  p(\vc = 1 | X, Y) \label{eq:em}
\end{align}
Since the above objective involves discrete latent variables $Y$, we propose to maximize Eq.(\ref{eq:em}) using the standard EM algorithm, especially its incremental, approximate variant. The target augmentation step in our approach is a sampled version of the E-step where the posterior samples are drawn with rejection sampling guided by the filter. The number of samples $K$ controls the quality of approximation to the posterior.\footnote{See Appendix \ref{more_molopt_exp} for details on the effect of sample size $K$.} 
The additional training step based on the augmented targets corresponds to a generalized M-step (though improvement is not guaranteed due to stochasticity). More precisely, let $P^{(t)}_\theta(Y|X)$ be the current translation model after $t$ epochs of augmentation training. In epoch $t+1$, the augmentation step first samples $C$ different candidates for each input $X$ using the old model $P^{(t)}$ parameterized by $\theta^{(t)}$, and then removes those which violate the constraint $\vc$; the remaining candidates are interpretable as samples from the current posterior $Q^{(t)}(Y|X) \propto P_{\theta}^{(t)}(Y|X) p(\vc = 1 | X, Y)$. As a result, the training step maximizes the EM auxiliary objective via stochastic gradient descent:
\begin{equation}
    J(\theta \;|\; \theta^{(t)}) = \mathbb{E}_X \left[  \sum_Y Q^{(t)}(Y|X)\, \log P_\theta(Y | X) \right]
\end{equation}
We train the model with multiple iterations and show empirically that model performance indeed keeps improving as we add more iterations, both in our main experiments as well as on a toy model in Appendix \ref{toy}. The EM approach is likely to converge to a different and better-performing translation model than the initial posterior calculation discussed in Equation \ref{initial_posterior}.
\section{Experiments}

We present experiments showcasing the effectiveness of our method, starting with conditional molecular design.

\subsection{Conditional Molecular Design}

The goal of conditional molecular design is to modify molecules to improve their chemical properties.
As illustrated in Figure~\ref{fig:molecule}, conditional molecular design is formulated as a graph-to-graph translation problem. 
The training data is a set of molecular pairs $\gD = \{(X_i,Y_i)\}$. $X$ is the input precursor and $Y$ is a similar molecule with improved properties. Each molecule is further labeled with its property score. 
Our method is well-suited to conditional molecular design because the target molecule is not unique: each precursor can be modified in many different ways to optimize its properties. Thus we can potentially discover several new targets per precursor during data augmentation.

\textbf{External Filter } The constraint contains two parts: 1) the chemical property of $Y$ must exceed a certain threshold $\beta$, and 2) the molecular similarity between $X$ and $Y$ must exceed a certain threshold $\delta$. The molecular similarity $\mathrm{sim}(X,Y)$ is defined as Tanimoto similarity on Morgan fingerprints~\citep{rogers2010extended}, which measures structural overlap between two molecules. 

In real-world settings, ground truth values of chemical properties are often evaluated through experimental assays, which are too expensive and time-consuming to run for stochastic iterative target augmentation. 
Therefore, we construct a proxy \textit{in silico} property predictor $F_1$ to approximate the true property evaluator $F_0$. 
To train this proxy predictor, we use the molecules in the training set and their labeled property values.
The proxy predictor $F_1$ is parameterized as a graph convolutional network and trained using the Chemprop package~\citep{yang2019learned}. During data augmentation, we use $F_1$ to filter out molecules whose predicted property score is under the threshold $\beta$.

\subsubsection{Experimental Setup}

We follow the evaluation setup of \citet{jin2018learning} for two conditional molecular design tasks: 

\begin{enumerate}[leftmargin=*,topsep=0pt,itemsep=0pt] 
    \item \textbf{QED Optimization}: The task is to improve the drug-likeness (QED) of a given compound $X$. The similarity constraint is $\mathrm{sim}(X,Y) \geq 0.4$ and the property constraint is $\mathrm{QED}(Y) \geq 0.9$, with $\mathrm{QED}(Y) \in [0,1]$ defined by the system of \citet{bickerton2012quantifying}.
    \item \textbf{DRD2 Optimization}: The task is to optimize biological activity against the dopamine type 2 receptor (DRD2). The similarity constraint is $\mathrm{sim}(X,Y) \geq 0.4$ and the property constraint is $\mathrm{DRD2}(Y) \geq 0.5$, where $\mathrm{DRD2}(Y) \in [0,1]$ is the predicted probability of biological activity given by the model from \citet{olivecrona2017molecular}.
\end{enumerate} 

We treat the output of the \textit{in silico} evaluators from \citet{bickerton2012quantifying} and \citet{olivecrona2017molecular} as ground truth, and we use them only during test-time evaluation to simulate a real-world scenario.\footnote{Although the Chemprop model we use in our filter is quite powerful, it fails to perfectly approximate the ground truth models for both QED and DRD2. The test set RMSE between our Chemprop model and the ground truth is 0.015 on the QED task and 0.059 on DRD2, where both properties range from 0 to 1.}

\begin{table*}[t] 
\centering
\begin{tabular}{@{}lcccc@{}}
\toprule
\textbf{Model} & \textbf{QED Succ.}  & \textbf{QED Div.} & \textbf{DRD2 Succ.} & \textbf{DRD2 Div.}       \\ \midrule
VSeq2Seq  & 58.5  & 0.331  & 75.9  & 0.176 \\
\textit{VSeq2Seq+} (Ours)  &  \textbf{89.0}  &  \textbf{0.470}  & \textbf{97.2}  & \textbf{0.361}  \\ 
\hdashline
\textit{VSeq2Seq+, semi-supervised} (Ours)* &  95.0 & 0.471  & 99.6 & 0.408  \\ 
\textit{VSeq2Seq+, transductive} (Ours)*  & 92.6   & 0.451 & 97.9 & 0.358 \\ \midrule
HierGNN  & 76.6 & 0.477  & 85.9  & 0.192 \\ 
\textit{HierGNN+} (Ours) & \textbf{93.1} & \textbf{0.514} & \textbf{97.6}  & \textbf{0.418} \\ \bottomrule

\end{tabular}
\caption{Performance of different models on QED and DRD2 conditional generation tasks. Italicized models with + are augmented by our algorithm. Best performance for each model architecture in bold, not including models that use additional unlabeled data. *Note that the semi-supervised and transductive settings for VSeq2Seq are not directly comparable to VSeq2Seq and VSeq2Seq+ due to using additional unlabeled data. However, they show that having access to such unlabeled inputs can substantially improve performance. But we emphasize that iterative target augmentation remains critical to performance in these settings: augmentation without an external filter instead decreases performance.}
\label{tab:results}
\end{table*}

\textbf{Evaluation Metrics.} During evaluation, we are interested both in the probability that the model finds a successful modification for a given molecule, as well as the diversity of the successful modifications when there are multiple. Thus we translate each molecule in the test set $Z=20$ times,\footnote{Our budget constraint $Z$ limits the number of accesses to the \textit{ground truth} evaluator, not the proxy predictor. In practice the ground truth evaluator is expensive while the proxy is cheap.} yielding candidate modifications $Y_1 \dots Y_Z$ (not necessarily distinct), and use the following two evaluation metrics:

\begin{enumerate}[leftmargin=*,topsep=0pt,itemsep=0pt] 
    \item \textit{Success}: The fraction of molecules $X$ for which \textit{any} of the outputs $Y_1 \dots Y_Z$ meet the required similarity and property constraints (specified previously for each task). This is our main metric.
    \item \textit{Diversity}: For each molecule $X$, we measure the average Tanimoto distance (defined as $1-\mathrm{sim}(Y_i,Y_j)$) between pairs within the set of successfully translated compounds among $Y_1 \dots Y_Z$. If there are one or fewer successful translations then the diversity is $0$. We average this quantity across all test precursors $X$.
\end{enumerate}

\begin{figure*}[t!]
    \centering
    \begin{subfigure}[t]{0.475\textwidth}
        \centering
        \includegraphics[height=1.6in]{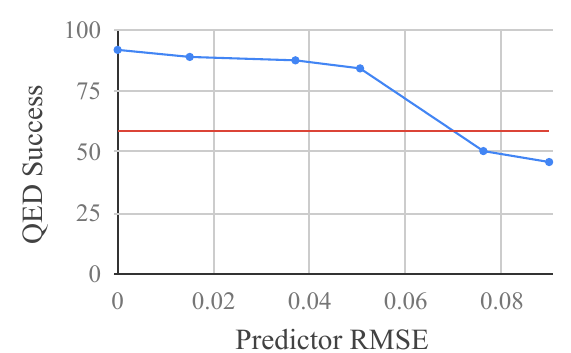}
    \end{subfigure}%
    \hfill
    \begin{subfigure}[t]{0.475\textwidth}
        \centering
        \includegraphics[height=1.6in]{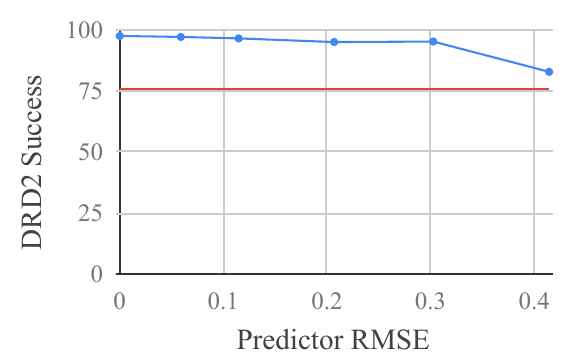}
    \end{subfigure}
    \vspace{-5pt}
    \caption{\textbf{Left}: QED test success rate vs. Chemprop predictor's RMSE with respect to ground truth. The red line shows the performance of the (unaugmented) VSeq2Seq baseline. \textbf{Right}: Same plot for DRD2. In each plot, the far left point with zero RMSE is obtained by reusing the ground truth predictor, while the second-from-left point is the Chemprop predictor we use to obtain our main results. Points further to the right are weaker predictors, simulating a scenario where the property is more difficult to model.}
    \label{fig:pred_rmse}
\end{figure*}

\begin{table*}[t] 
\centering
\begin{tabular}{@{}lccccccc@{}}
\toprule
\textbf{Model} & \textbf{Train-Aug} & \textbf{Train+} & \textbf{Test+} & \textbf{QED Succ.}  & \textbf{QED Div.} & \textbf{DRD2 Succ.} & \textbf{DRD2 Div.} \\ \midrule
VSeq2Seq         & {\color{red}\xmark } & {\color{red}\xmark }               & {\color{red}\xmark }                        & 58.5        & 0.331         & 75.9         & 0.176          \\
\textit{VSeq2Seq(test)} & {\color{red}\xmark } &{\color{red}\xmark }               & {\color{green}\cmark }                       & 77.4        & \textbf{0.471}         & 87.2         & 0.200            \\
\textit{VSeq2Seq(train)} & {\color{green}\cmark }& {\color{green}\cmark }               & {\color{red}\xmark }                        & 81.8        & 0.430          & 92.2         & 0.321          \\
\textit{VSeq2Seq+}        & {\color{green}\cmark }& {\color{green}\cmark }               & {\color{green}\cmark }                       & \textbf{89.0}          & \textbf{0.470}          & \textbf{97.2}         & \textbf{0.361}         \\
\textit{VSeq2Seq(no-filter)}        & {\color{green}\cmark }& {\color{red}\xmark }               & {\color{red}\xmark }                       & 47.5          & 0.297          & 51.0         & 0.185         \\\bottomrule
\end{tabular}
\caption{Ablation analysis of filtering at training and test time. ``Train-Aug'' indicates a model whose training process uses self-generated candidates to augment the data, while ``Train+'' is a model that additionally filters these candidates using the proxy according to our framework. ``Test+'' indicates a model that filters outputs at prediction time using the learned proxy predictor. We emphasize that the ground truth predictor is used only for final evaluation. The evaluation for VSeq2Seq(no-filter) is conducted after 10 augmentation epochs, as the best validation set performance only decreases over the course of training.}
\label{tab:filter}
\end{table*}

\textbf{Models and Baselines.} We consider the following two model architectures from \citet{jin2019multi} to show that our algorithm is not tied to specific neural architectures.
\begin{enumerate}[leftmargin=*,topsep=0pt,itemsep=0pt] 
    \item VSeq2Seq, a sequence-to-sequence translation model generating molecules by their SMILES string \citep{weininger1988smiles}.
    \item HierGNN, a hierarchical graph-to-graph architecture that achieves state-of-the-art performance on the QED and DRD2 tasks, outperforming VSeq2Seq by a wide margin.
\end{enumerate}

We apply our iterative augmentation procedure to the above two models, generating up to $K=4$ new targets per precursor in each augmentation epoch. Additionally, we evaluate our augmentation of VSeq2Seq in a transductive setting, as well as in a semi-supervised setting where we provide 100K additional source-side precursors from the ZINC database~\citep{sterling2015zinc}. Full hyperparameters are provided in Appendix \ref{hyperparams}.

\subsubsection{Results}

As shown in Table \ref{tab:results}, our iterative augmentation paradigm significantly improves the performance of VSeq2Seq and HierGNN. On both datasets, the translation success rate increases by over 10\% in absolute terms for both models. In fact, VSeq2Seq+, our augmentation of the simple VSeq2Seq model, outperforms the non-augmented version of HierGNN. This result strongly confirms our hypothesis about the inherent challenge of learning translation models in data-sparse scenarios. Moreover, we find that adding more precursors during data augmentation further improves the VSeq2Seq model. On the QED dataset, the translation success rate improves from 89.0\% to 92.6\% by just adding test set molecules as precursors (VSeq2Seq+, transductive). When instead adding 100K precursors from the external ZINC database, the performance further increases to 95.0\% (VSeq2Seq+, semi-supervised). We observe similar improvements for the DRD2 task as well. Beyond accuracy gain, our augmentation strategy also improves the diversity of generated molecules. For instance, on the DRD2 task, our approach yields a 100\% relative gain in output diversity. 

These improvements over the baselines are perhaps unsurprising when considering the much greater amount of augmented ``data" pairs seen by our augmented model. For example, VSeq2Seq+ has seen over 20 times as much ``data" as the base model by the end of training on the QED task (Figure 4). 

\begin{center}\label{fig:qed_seen_data}
    \includegraphics[height=1.6in]{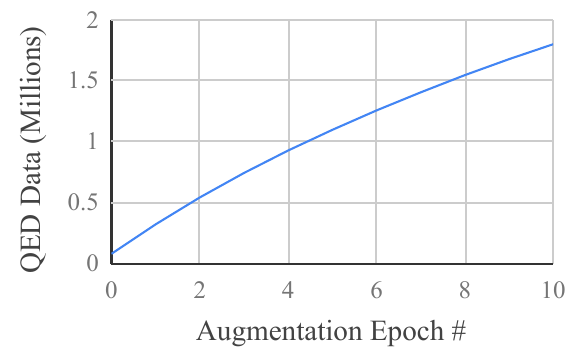}
    \captionof{figure}{Cumulative number of unique training pairs seen by VSeq2Seq+ model after each augmentation epoch, on QED task.}
\end{center}

\textbf{Importance of Property Predictor } 
Although the property predictor used in data augmentation differs from the ground truth property evaluator used at test time, the difference in evaluators does not derail the overall training process.
Here we analyze the influence of the quality of the property predictor used in data augmentation. Specifically, we rerun our experiments using less accurate proxy predictors for our external
filter. We obtain these weakened predictors by undertraining Chemprop and decreasing its hidden dimension. For comparison, we also report results with the oracle property predictor which is the ground truth evaluator. 

As shown in Figure~\ref{fig:pred_rmse}, on the DRD2 dataset we can maintain strong performance despite using predictors that deviate significantly from the ground truth. 
This implies that our framework can potentially be applied to other properties that are harder to predict.
On the QED dataset, our method is less tolerant of inaccurate property prediction because the property constraint is much tighter --- it requires the QED score of an output $Y$ to be in the range [0.9, 1.0].

\textbf{Importance of External Filtering} 
Our full model VSeq2Seq+ uses the external filter during both training and testing.
We further experiment with Vseq2seq(test), a version of our model trained without data augmentation but which uses the external filter to remove invalid outputs at test time. As shown in Table \ref{tab:filter}, VSeq2Seq(test) performs significantly worse than our full model trained under data augmentation. Similarly, a model VSeq2Seq(train) trained with data augmentation but without prediction time filtering also performs much worse than the full model. 

We also run an augmentation-only version of the model without an external filter. This model (referred to as VSeq2Seq(no-filter) in Table \ref{tab:filter}) augments the data in each epoch by simply using the first $K$ distinct candidate translations for each training precursor $X$, without using the external filter at all. We additionally provide this model with the 100K unlabeled precursors from the semi-supervised setting. Nevertheless, we find that during augmentation, this model's performance steadily declines from that of the bootstrapped prior. Thus the external filter is necessary to prevent poor targets from leading the model training astray. 


\subsection{Unconditional Molecular Design}

In unconditional molecular design, we learn a distribution over molecules with desired properties. The setup is similar to the conditional case, and we reuse the same QED and DRD2 datasets. However, as there is no input in the unconditional case, we drop the precursors $X$ and use only the set of targets $Y$ as our training data. Additionally, we drop the similarity component from our external filter; we now require only that each generated molecule has sufficiently high property score. We use the same property thresholds for the QED and DRD2 tasks as in the conditional case. 

\begin{table*}[t]
\centering
\begin{tabular}{@{}lcccc@{}}
\toprule
\textbf{Model} & \textbf{QED Succ.}  & \textbf{QED Uniq.} & \textbf{DRD2 Succ.} &\textbf{DRD2 Uniq.} \\ \midrule
VSeq                                & 62.4     & 0.499         & 51.4       & 0.221          \\
\textit{VSeq+} (Ours)                        & \textbf{95.8}       & \textbf{0.957}          & \textbf{92.8}        & \textbf{0.927}          \\\midrule
REINVENT                               & 61.9        & 0.610          & \textbf{92.2}    & 0.686            \\\bottomrule
\end{tabular}
\caption{Performance of different models on QED and DRD2 unconditional generation tasks. VSeq+ is our full augmented model.}
\label{tab:unc_results}
\end{table*}

\begin{figure*}[t!]
    \centering
    \begin{subfigure}[t]{0.475\textwidth}
        \centering
        \includegraphics[height=1.6in]{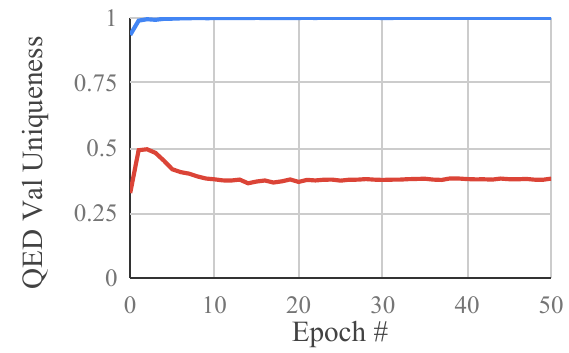}
    \end{subfigure}%
    \hfill
    \begin{subfigure}[t]{0.475\textwidth}
        \centering
        \includegraphics[height=1.6in]{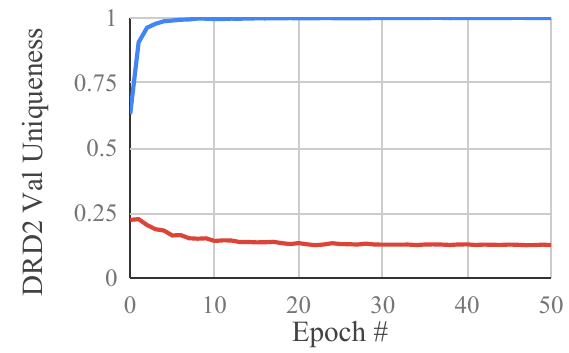}
    \end{subfigure}
    \vspace{-5pt}
    \caption{\textbf{Left}: Epoch number vs. uniqueness, evaluated with the Chemprop proxy predictor, for VSeq-based models on QED dataset. VSeq+ and VSeq in blue and red respectively. \textbf{Right}: Same plot for DRD2. VSeq+ is trained without iterative target augmentation for the initial epoch 0, and trained with augmentation thereafter.}
    \label{fig:uniqueness}
\end{figure*}

\textbf{Evaluation Metrics.} We modify our metrics for the unconditional case:

\begin{enumerate}[leftmargin=*,topsep=0pt,itemsep=0pt] 
    \item \textit{Success}: The fraction of sampled molecules $Y$ above the property score threshold. 
    \item \textit{Uniqueness}: The number of unique molecules generated in $20000$ samples passing the property score threshold, as a fraction of $20000$. This is our main metric. 
\end{enumerate}

In the unconditional case, a model can achieve perfect success and high pairwise diversity simply by memorizing a small number of molecules with high property score. Therefore, uniqueness is our main metric in the unconditional setting, as a diverse distribution of molecules with high property scores is necessary to achieve high uniqueness.

\textbf{Models and Baselines.} We consider two baselines:

\begin{enumerate} [leftmargin=*,topsep=0pt,itemsep=0pt] 
    \item A modified version of VSeq2Seq which simply drops the input and corresponding attention layers; the resulting model is essentially a variational autoencoder~\citep{kingma2013auto}. We refer to this model as VSeq. 
    \item REINVENT, a sequence-based model from \citet{olivecrona2017molecular} which uses the external property scorer to fine-tune the model via reinforcement learning. This can be viewed as an alternate method of leveraging the external filter. We note that although \citet{olivecrona2017molecular} also originally evaluated on the DRD2 property, our setup is more challenging: we allow significantly less training data for bootstrapping, and prohibit the use of the ground truth predictor before test time. 
\end{enumerate}

REINVENT and our augmented model VSeq+ (obtained by augmenting VSeq) are trained to convergence. For VSeq, whose uniqueness score decreases with prolonged training, we choose the checkpoint maximizing uniqueness under the Chemprop proxy predictor. Although the VSeq and REINVENT architectures differ slightly, we match the number of trainable parameters. We provide full hyperparameters and ablations in Appendices \ref{hyperparams} and \ref{uc_ablations} respectively.


\subsubsection{Results}\label{uc_results}


As shown in Table \ref{tab:unc_results}, our iterative augmentation scheme significantly improves the performance of VSeq, especially in uniqueness. In fact, uniqueness steadily decreases over time for the VSeq baseline as it overfits the training data (Figure \ref{fig:uniqueness}). On the other hand, our augmented model VSeq+ sees a steady increase in uniqueness over time.

Moreover, our iterative augmentation scheme outperforms the REINVENT baseline on both tasks by over 0.2 in absolute terms. Especially on the QED task, the REINVENT algorithm struggles to generate high-property molecules consistently, performing comparably to the unaugmented VSeq baseline in success rate. Additionally, we observed that the REINVENT model is sometimes unstable on our DRD2 task, where the initial training dataset is smaller. Meanwhile, VSeq+ showed consistently strong performance on both tasks. Overall our experiments in this unconditional setting indicate that stochastic iterative target augmentation, at least in certain scenarios, is capable of leveraging the external property signal more effectively than an RL method. 

\begin{figure*}[]
  \centering
  \includegraphics[width=\linewidth]{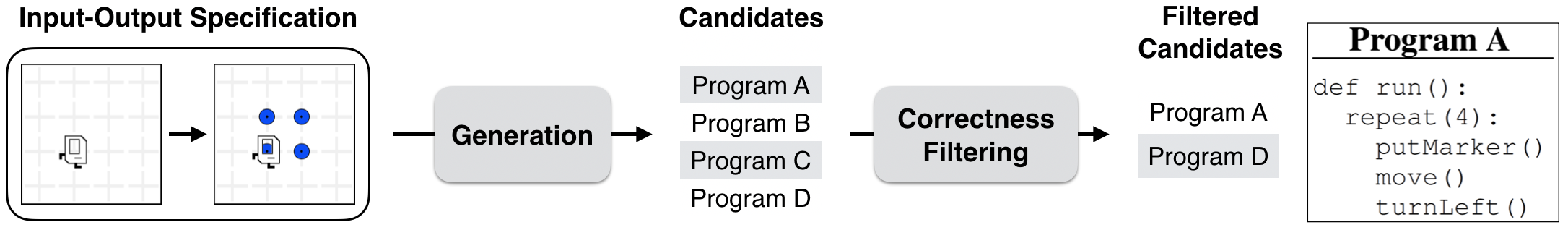}
  \caption{Illustration of our data generation process in the program synthesis setting. Given an input-output specification, we first use our generation model to generate candidate programs, and then select correct programs using our external filter. Images of input-output specification and the program A are from \citet{bunel2018leveraging}.}
  \label{fig:train_filter_karel}
\end{figure*}

\subsection{Program Synthesis Experiments}\label{program_synth}

Finally, we present additional experiments using the conditional version of our method in the program synthesis domain, demonstrating its generalizability across domains. 
Program synthesis is the task of generating a program (using domain-specific language) based on given input-output specifications \citep{bunel2018leveraging,gulwani2011automating,devlin2017neural}. 
That is, the source is a set of input-output specifications for the program, and the target is a program that passes all test cases. Our method is suitable for this task because the target program is not unique. Multiple programs may be consistent with the given input-output specifications.
\newline
\newline

\textbf{External Filter} The external filter is straightforward for this task: we simply check whether the generated output passes all test cases. Note that at evaluation time, each instance contains extra held-out input-output test cases; the program must pass these in addition to the given test cases to be considered correct. When we perform prediction time filtering, we do not use held-out test cases in our filter.

\subsubsection{Experimental Setup}

Our task is based on the educational Karel programming language~\citep{Pattis:1981:KRG:539521} used for evaluation in \citet{bunel2018leveraging} and \citet{chen2018execution}. Commands in the Karel language guide a robot's actions in a 2D grid, and may include for loops, while loops, and conditionals. Figure \ref{fig:train_filter_karel} contains an example. We follow the experiment setup of \citet{bunel2018leveraging}. 

\textbf{Evaluation Metrics.} The evaluation metric is top-1 generalization. This metric measures how often the model can generate a program that passes the input-output test cases on the test set. At test time, we use our model to generate up to $L$ candidate programs and select the first one to pass the input-output specifications (not including held-out test cases). 

\textbf{Models and Baselines.} Our main baseline is the MLE baseline from \citet{bunel2018leveraging}. This model consists of a CNN encoder for the input-output grids and an LSTM decoder along with a hand-coded syntax checker. 
It is trained to maximize the likelihood of the provided target program.
Our model is the augmentation of this MLE baseline by our iterative target augmentation framework. As with molecular design, we generate up to $K=4$ new targets per precursor during each augmentation step. 
Additionally, we compare against the best model from \citet{bunel2018leveraging}, which finetunes the same MLE architecture using an RL method with beam search to estimate gradients.\footnote{More recently, \citet{chen2018execution} achieved state-of-the-art performance on the same Karel task, with top-1 generalization accuracy of 92\%. They use a different architecture highly specialized for program synthesis as well as a specialized ensemble method. Thus their results are not directly comparable to our results in this paper for the MLE architecture.} We use the same hyperparameters as the original MLE baseline; see Appendix~\ref{hyperparams} for details.

\subsubsection{Results}

\begin{table}[]
\centering
\begin{tabular}{@{}p{4cm}c@{}}
\toprule
\textbf{Model} & \textbf{Top-1} \\ \midrule
MLE  & 71.91 \\ 
MLE + RL + Beam Search & 77.12  \\ 
\textit{MLE+} (Ours) &  \textbf{85.02} \\ \bottomrule 
\end{tabular}
\captionof{table}{Model performance measured by top-1 generalization accuracy on Karel program synthesis task. MLE+ is our augmented version of the MLE model~\citep{bunel2018leveraging}, while MLE + RL + Beam Search is their reinforcement learning method applied to the same architecture.}
\label{tab:code_results}

\end{table}

Table \ref{tab:code_results} shows the performance of our model in comparison to previous work. Our model (MLE+) outperforms the base MLE model in \citet{bunel2018leveraging} model by a wide margin.
Moreover, our model outperforms the best reinforcement learning model (RL + Beam Search) in \citet{bunel2018leveraging}, which was trained to directly maximize the generalization metric. This demonstrates the efficacy of our approach in the program synthesis domain. Since our method is complementary to architectural improvements, we hypothesize that other techniques, such as execution based synthesis~\citep{chen2018execution}, can benefit from our approach as well.
\section{Related Work}

\textbf{Molecular Design } 
Several previous works explore molecular design using different architectures. 
\citet{segler2017generating,kusner2017grammar,gomez2016automatic,kang2018conditional} adopt generative modeling approaches for molecular design. \citet{you2018graph,popova2018deep,olivecrona2017molecular} use reinforcement learning methods for this task. \citet{jin2019multi,jin2018learning} formulate this problem as graph-to-graph translation and significantly outperform previous methods in the conditional setting. However, their performance remains imperfect due to the limited size of given training sets.

On the other hand, recent advances in graph convolutional networks \citep{duvenaud2015convolutional,gilmer2017neural} have provided effective solutions for the related problem of property prediction. 
Our work leverages strong property prediction models to improve the performance of generative models for molecular design, by checking whether generated molecules have desired chemical properties and augmenting the training set with molecules passing the property filter. 

\textbf{Program Synthesis } 
When correctness in program synthesis is defined by input-output test cases~\citep{bunel2018leveraging,gulwani2011automating,devlin2017neural}, one can check a generated program's correctness by simply executing it on each input and verifying its output. Indeed, \citet{zhang2018leveraging,chen2018execution} use this idea in their respective decoding procedures, while also using structural constraints on valid programs. We leverage this ability to check correctness during training time data augmentation as well. 

\textbf{Reward-guided Generation }
Recent work has proposed to incorporate rewards (e.g., properties) into generative models. 
In machine translation, \citet{norouzi2016reward} propose reward augmented maximum likelihood, which samples new targets from a \emph{stationary} exponentiated payoff distribution centered at a ground truth target based on edit distance. Their approach is only viable when ground truth targets are given. In the case of molecular design, the number of ground truth targets is very limited. Our approach, based on stochastic EM, samples new targets from a learned non-stationary distribution which is not tied to any ground truth.

\citet{jaques2017sequence} use reinforcement learning to impose task-specific rewards for sequence generation, while \citet{brookes2019conditioning} propose an adaptive sampling approach which generates additional targets based on parametric conditional density estimation. In contrast to these two approaches, our method is based on maximum likelihood and stochastic EM; \citet{brookes2019view} explore additional theoretical connections. 


\textbf{Semi-supervised Learning } 
Our method is related to various approaches to semi-supervised learning in different domains. 
In chemistry, \citet{hu2019pre} and \citet{sun2019infograph} demonstrate pre-training approaches which use unlabeled molecules to learn initial representations for property prediction models. 
Our method instead tackles the problem of molecular generation, addressing the problem of limited data by generating additional data via a self-training technique. 
In machine translation, back-translation \citep{sennrich2015improving,edunov2018understanding} creates additional translation pairs by using a backward translation system to translate unlabeled sentences from a target language into a source language. In contrast, our method works in the forward direction because many translation tasks are not symmetric. 

In image and text classification, data augmentation and label guessing \citep{lee2013pseudo,berthelot2019mixmatch,xie2019unsupervised} are commonly applied to obtain artificial labels for unlabeled data. Rather than generating new source-target pairs by augmenting the source side, we augment the target side.
In syntactic parsing, our method is closely related to self-training~\citep{mcclosky2006effective}. They generate new parse trees from unlabeled sentences by applying an existing parser followed by a reranker, and then treat the resulting parse trees as new training targets. 
However, their method is not iterative, and their reranker is explicitly trained to operate over the top $k$ outputs of the parser; in contrast, our filter is independent of the generative model. 
In addition we show that our approach, which can be viewed as iteratively combining reranking and self-training, is theoretically motivated and can improve the performance of highly complex neural models.
Co-training~\citep{blum1998combining} and tri-training~\citep{zhou2005tri,charniak2016parsing} also augment a parsing dataset by adding targets on which multiple baseline models agree. Instead of using multiple learners, our method uses task-specific constraints to select correct outputs.

\section{Conclusion}

In this work, we have presented a stochastic iterative target augmentation framework for molecular design. Our approach is theoretically motivated, and we demonstrate strong empirical results in both the conditional and unconditional molecular design settings, significantly outperforming baseline models in each case. Moreover, we find that stochastic iterative target augmentation is complementary to architectural improvements, and that its effect can be quite robust to the external filter's quality. Finally, in principle our approach is applicable to other domains as well.

\section*{Acknowledgements}
We thank Guang-He Lee, Yujia Bao, Rachel Wu, Benson Chen, Mitchell Stern, Nikita Kitaev, Ruiqi Zhong, Rudy Corona, Eric Wallace, Kevin Lin, Daniel Fried, Dan Klein, and our anonymous reviewers for their helpful comments and feedback which helped us to greatly improve the paper. 

\bibliographystyle{icml2020}
\bibliography{references}

\begin{thebibliography}{43}
\providecommand{\natexlab}[1]{#1}
\providecommand{\url}[1]{\texttt{#1}}
\expandafter\ifx\csname urlstyle\endcsname\relax
  \providecommand{\doi}[1]{doi: #1}\else
  \providecommand{\doi}{doi: \begingroup \urlstyle{rm}\Url}\fi

\bibitem[Berthelot et~al.(2019)Berthelot, Carlini, Goodfellow, Papernot,
  Oliver, and Raffel]{berthelot2019mixmatch}
Berthelot, D., Carlini, N., Goodfellow, I., Papernot, N., Oliver, A., and
  Raffel, C.
\newblock Mixmatch: A holistic approach to semi-supervised learning.
\newblock \emph{arXiv preprint arXiv:1905.02249}, 2019.

\bibitem[Bickerton et~al.(2012)Bickerton, Paolini, Besnard, Muresan, and
  Hopkins]{bickerton2012quantifying}
Bickerton, G.~R., Paolini, G.~V., Besnard, J., Muresan, S., and Hopkins, A.~L.
\newblock Quantifying the chemical beauty of drugs.
\newblock \emph{Nature chemistry}, 4\penalty0 (2):\penalty0 90, 2012.

\bibitem[Blum \& Mitchell(1998)Blum and Mitchell]{blum1998combining}
Blum, A. and Mitchell, T.
\newblock Combining labeled and unlabeled data with co-training.
\newblock In \emph{Proceedings of the eleventh annual conference on
  Computational learning theory}, pp.\  92--100. Citeseer, 1998.

\bibitem[Brookes et~al.(2019{\natexlab{a}})Brookes, Busia, Fannjiang, Murphy,
  and Listgarten]{brookes2019view}
Brookes, D.~H., Busia, A., Fannjiang, C., Murphy, K., and Listgarten, J.
\newblock A view of estimation of distribution algorithms through the lens of
  expectation-maximization.
\newblock \emph{arXiv preprint arXiv:1905.10474}, 2019{\natexlab{a}}.

\bibitem[Brookes et~al.(2019{\natexlab{b}})Brookes, Park, and
  Listgarten]{brookes2019conditioning}
Brookes, D.~H., Park, H., and Listgarten, J.
\newblock Conditioning by adaptive sampling for robust design.
\newblock \emph{arXiv preprint arXiv:1901.10060}, 2019{\natexlab{b}}.

\bibitem[Bunel et~al.(2018)Bunel, Hausknecht, Devlin, Singh, and
  Kohli]{bunel2018leveraging}
Bunel, R., Hausknecht, M., Devlin, J., Singh, R., and Kohli, P.
\newblock Leveraging grammar and reinforcement learning for neural program
  synthesis.
\newblock \emph{arXiv preprint arXiv:1805.04276}, 2018.

\bibitem[Celeux et~al.(1996)Celeux, Chauveau, and
  Diebolt]{celeux1996stochastic}
Celeux, G., Chauveau, D., and Diebolt, J.
\newblock Stochastic versions of the em algorithm: an experimental study in the
  mixture case.
\newblock \emph{Journal of statistical computation and simulation}, 55\penalty0
  (4):\penalty0 287--314, 1996.

\bibitem[Charniak et~al.(2016)]{charniak2016parsing}
Charniak, E. et~al.
\newblock Parsing as language modeling.
\newblock In \emph{Proceedings of the 2016 Conference on Empirical Methods in
  Natural Language Processing}, pp.\  2331--2336, 2016.

\bibitem[Chen et~al.(2019)Chen, Liu, and Song]{chen2018execution}
Chen, X., Liu, C., and Song, D.
\newblock Execution-guided neural program synthesis.
\newblock \emph{International Conference on Learning Representations}, 2019.

\bibitem[Devlin et~al.(2017)Devlin, Bunel, Singh, Hausknecht, and
  Kohli]{devlin2017neural}
Devlin, J., Bunel, R.~R., Singh, R., Hausknecht, M., and Kohli, P.
\newblock Neural program meta-induction.
\newblock In \emph{Advances in Neural Information Processing Systems}, pp.\
  2080--2088, 2017.

\bibitem[Duvenaud et~al.(2015)Duvenaud, Maclaurin, Iparraguirre, Bombarell,
  Hirzel, Aspuru-Guzik, and Adams]{duvenaud2015convolutional}
Duvenaud, D.~K., Maclaurin, D., Iparraguirre, J., Bombarell, R., Hirzel, T.,
  Aspuru-Guzik, A., and Adams, R.~P.
\newblock Convolutional networks on graphs for learning molecular fingerprints.
\newblock \emph{Advances in Neural Information Processing Systems}, pp.\
  2224--2232, 2015.

\bibitem[Edunov et~al.(2018)Edunov, Ott, Auli, and
  Grangier]{edunov2018understanding}
Edunov, S., Ott, M., Auli, M., and Grangier, D.
\newblock Understanding back-translation at scale.
\newblock \emph{arXiv preprint arXiv:1808.09381}, 2018.

\bibitem[Gilmer et~al.(2017)Gilmer, Schoenholz, Riley, Vinyals, and
  Dahl]{gilmer2017neural}
Gilmer, J., Schoenholz, S.~S., Riley, P.~F., Vinyals, O., and Dahl, G.~E.
\newblock Neural message passing for quantum chemistry.
\newblock \emph{Proceedings of the 34th International Conference on Machine
  Learning}, 2017.

\bibitem[G{\'o}mez-Bombarelli et~al.(2018)G{\'o}mez-Bombarelli, Wei, Duvenaud,
  Hern{\'a}ndez-Lobato, S{\'a}nchez-Lengeling, Sheberla, Aguilera-Iparraguirre,
  Hirzel, Adams, and Aspuru-Guzik]{gomez2016automatic}
G{\'o}mez-Bombarelli, R., Wei, J.~N., Duvenaud, D., Hern{\'a}ndez-Lobato,
  J.~M., S{\'a}nchez-Lengeling, B., Sheberla, D., Aguilera-Iparraguirre, J.,
  Hirzel, T.~D., Adams, R.~P., and Aspuru-Guzik, A.
\newblock Automatic chemical design using a data-driven continuous
  representation of molecules.
\newblock \emph{ACS Central Science}, 2018.
\newblock \doi{10.1021/acscentsci.7b00572}.

\bibitem[Gulwani(2011)]{gulwani2011automating}
Gulwani, S.
\newblock Automating string processing in spreadsheets using input-output
  examples.
\newblock In \emph{ACM Sigplan Notices}, volume~46, pp.\  317--330. ACM, 2011.

\bibitem[Hastings(1970)]{hastings1970monte}
Hastings, W.~K.
\newblock Monte carlo sampling methods using markov chains and their
  applications.
\newblock 1970.

\bibitem[Heusel et~al.(2017)Heusel, Ramsauer, Unterthiner, Nessler, and
  Hochreiter]{heusel2017gans}
Heusel, M., Ramsauer, H., Unterthiner, T., Nessler, B., and Hochreiter, S.
\newblock Gans trained by a two time-scale update rule converge to a local nash
  equilibrium.
\newblock In \emph{Advances in neural information processing systems}, pp.\
  6626--6637, 2017.

\bibitem[Hu et~al.(2019)Hu, Liu, Gomes, Zitnik, Liang, Pande, and
  Leskovec]{hu2019pre}
Hu, W., Liu, B., Gomes, J., Zitnik, M., Liang, P., Pande, V., and Leskovec, J.
\newblock Pre-training graph neural networks.
\newblock \emph{arXiv preprint arXiv:1905.12265}, 2019.

\bibitem[Jaques et~al.(2017)Jaques, Gu, Bahdanau, Hern{\'a}ndez-Lobato, Turner,
  and Eck]{jaques2017sequence}
Jaques, N., Gu, S., Bahdanau, D., Hern{\'a}ndez-Lobato, J.~M., Turner, R.~E.,
  and Eck, D.
\newblock Sequence tutor: Conservative fine-tuning of sequence generation
  models with kl-control.
\newblock In \emph{Proceedings of the 34th International Conference on Machine
  Learning-Volume 70}, pp.\  1645--1654. JMLR. org, 2017.

\bibitem[Jin et~al.(2019{\natexlab{a}})Jin, Barzilay, and
  Jaakkola]{jin2019multi}
Jin, W., Barzilay, R., and Jaakkola, T.
\newblock Multi-resolution autoregressive graph-to-graph translation for
  molecules.
\newblock \emph{arXiv preprint arXiv:1907.11223}, 2019{\natexlab{a}}.

\bibitem[Jin et~al.(2019{\natexlab{b}})Jin, Yang, Barzilay, and
  Jaakkola]{jin2018learning}
Jin, W., Yang, K., Barzilay, R., and Jaakkola, T.
\newblock Learning multimodal graph-to-graph translation for molecular
  optimization.
\newblock \emph{International Conference on Learning Representation},
  2019{\natexlab{b}}.

\bibitem[Kang \& Cho(2018)Kang and Cho]{kang2018conditional}
Kang, S. and Cho, K.
\newblock Conditional molecular design with deep generative models.
\newblock \emph{Journal of chemical information and modeling}, 59\penalty0
  (1):\penalty0 43--52, 2018.

\bibitem[Kingma \& Welling(2013)Kingma and Welling]{kingma2013auto}
Kingma, D.~P. and Welling, M.
\newblock Auto-encoding variational bayes.
\newblock \emph{arXiv preprint arXiv:1312.6114}, 2013.

\bibitem[Kusner et~al.(2017)Kusner, Paige, and
  Hern{\'a}ndez-Lobato]{kusner2017grammar}
Kusner, M.~J., Paige, B., and Hern{\'a}ndez-Lobato, J.~M.
\newblock Grammar variational autoencoder.
\newblock \emph{arXiv preprint arXiv:1703.01925}, 2017.

\bibitem[Lee(2013)]{lee2013pseudo}
Lee, D.-H.
\newblock Pseudo-label: The simple and efficient semi-supervised learning
  method for deep neural networks.
\newblock In \emph{Workshop on challenges in representation learning, ICML},
  volume~3, pp.\ ~2, 2013.

\bibitem[McClosky et~al.(2006)McClosky, Charniak, and
  Johnson]{mcclosky2006effective}
McClosky, D., Charniak, E., and Johnson, M.
\newblock Effective self-training for parsing.
\newblock In \emph{Proceedings of the main conference on human language
  technology conference of the North American Chapter of the Association of
  Computational Linguistics}, pp.\  152--159. Association for Computational
  Linguistics, 2006.

\bibitem[Norouzi et~al.(2016)Norouzi, Bengio, Jaitly, Schuster, Wu, Schuurmans,
  et~al.]{norouzi2016reward}
Norouzi, M., Bengio, S., Jaitly, N., Schuster, M., Wu, Y., Schuurmans, D.,
  et~al.
\newblock Reward augmented maximum likelihood for neural structured prediction.
\newblock In \emph{Advances In Neural Information Processing Systems}, pp.\
  1723--1731, 2016.

\bibitem[Olivecrona et~al.(2017)Olivecrona, Blaschke, Engkvist, and
  Chen]{olivecrona2017molecular}
Olivecrona, M., Blaschke, T., Engkvist, O., and Chen, H.
\newblock Molecular de-novo design through deep reinforcement learning.
\newblock \emph{Journal of cheminformatics}, 9\penalty0 (1):\penalty0 48, 2017.

\bibitem[Paszke et~al.(2017)Paszke, Gross, Chintala, Chanan, Yang, DeVito, Lin,
  Desmaison, Antiga, and Lerer]{paszke2017automatic}
Paszke, A., Gross, S., Chintala, S., Chanan, G., Yang, E., DeVito, Z., Lin, Z.,
  Desmaison, A., Antiga, L., and Lerer, A.
\newblock Automatic differentiation in pytorch.
\newblock 2017.

\bibitem[Pattis(1981)]{Pattis:1981:KRG:539521}
Pattis, R.~E.
\newblock \emph{Karel the Robot: A Gentle Introduction to the Art of
  Programming}.
\newblock John Wiley \& Sons, Inc., New York, NY, USA, 1st edition, 1981.
\newblock ISBN 0471089281.

\bibitem[Popova et~al.(2018)Popova, Isayev, and Tropsha]{popova2018deep}
Popova, M., Isayev, O., and Tropsha, A.
\newblock Deep reinforcement learning for de novo drug design.
\newblock \emph{Science advances}, 4\penalty0 (7):\penalty0 eaap7885, 2018.

\bibitem[Preuer et~al.(2018)Preuer, Renz, Unterthiner, Hochreiter, and
  Klambauer]{preuer2018frechet}
Preuer, K., Renz, P., Unterthiner, T., Hochreiter, S., and Klambauer, G.
\newblock Fr{\'e}chet chemnet distance: a metric for generative models for
  molecules in drug discovery.
\newblock \emph{Journal of chemical information and modeling}, 58\penalty0
  (9):\penalty0 1736--1741, 2018.

\bibitem[Rogers \& Hahn(2010)Rogers and Hahn]{rogers2010extended}
Rogers, D. and Hahn, M.
\newblock Extended-connectivity fingerprints.
\newblock \emph{J. Chem. Inf. Model.}, 50\penalty0 (5):\penalty0 742--754,
  2010.

\bibitem[Segler et~al.(2017)Segler, Kogej, Tyrchan, and
  Waller]{segler2017generating}
Segler, M.~H., Kogej, T., Tyrchan, C., and Waller, M.~P.
\newblock Generating focussed molecule libraries for drug discovery with
  recurrent neural networks.
\newblock \emph{arXiv preprint arXiv:1701.01329}, 2017.

\bibitem[Sennrich et~al.(2015)Sennrich, Haddow, and
  Birch]{sennrich2015improving}
Sennrich, R., Haddow, B., and Birch, A.
\newblock Improving neural machine translation models with monolingual data.
\newblock \emph{arXiv preprint arXiv:1511.06709}, 2015.

\bibitem[Sterling \& Irwin(2015)Sterling and Irwin]{sterling2015zinc}
Sterling, T. and Irwin, J.~J.
\newblock Zinc 15--ligand discovery for everyone.
\newblock \emph{Journal of chemical information and modeling}, 55\penalty0
  (11):\penalty0 2324--2337, 2015.

\bibitem[Sun et~al.(2019)Sun, Hoffmann, and Tang]{sun2019infograph}
Sun, F.-Y., Hoffmann, J., and Tang, J.
\newblock Infograph: Unsupervised and semi-supervised graph-level
  representation learning via mutual information maximization.
\newblock \emph{arXiv preprint arXiv:1908.01000}, 2019.

\bibitem[Weininger(1988)]{weininger1988smiles}
Weininger, D.
\newblock Smiles, a chemical language and information system. 1. introduction
  to methodology and encoding rules.
\newblock \emph{J. Chem. Inf. Model.}, 28\penalty0 (1):\penalty0 31--36, 1988.

\bibitem[Xie et~al.(2019)Xie, Dai, Hovy, Luong, and Le]{xie2019unsupervised}
Xie, Q., Dai, Z., Hovy, E., Luong, M.-T., and Le, Q.~V.
\newblock Unsupervised data augmentation.
\newblock \emph{arXiv preprint arXiv:1904.12848}, 2019.

\bibitem[Yang et~al.(2019)Yang, Swanson, Jin, Coley, Eiden, Gao, Guzman-Perez,
  Hopper, Kelley, Mathea, et~al.]{yang2019learned}
Yang, K., Swanson, K., Jin, W., Coley, C.~W., Eiden, P., Gao, H., Guzman-Perez,
  A., Hopper, T., Kelley, B., Mathea, M., et~al.
\newblock Analyzing learned molecular representations for property prediction.
\newblock \emph{Journal of chemical information and modeling}, 2019.

\bibitem[You et~al.(2018)You, Liu, Ying, Pande, and Leskovec]{you2018graph}
You, J., Liu, B., Ying, Z., Pande, V., and Leskovec, J.
\newblock Graph convolutional policy network for goal-directed molecular graph
  generation.
\newblock In \emph{Advances in Neural Information Processing Systems}, pp.\
  6410--6421, 2018.

\bibitem[Zhang et~al.(2018)Zhang, Rosenblatt, Fetaya, Liao, Byrd, Urtasun, and
  Zemel]{zhang2018leveraging}
Zhang, L., Rosenblatt, G., Fetaya, E., Liao, R., Byrd, W.~E., Urtasun, R., and
  Zemel, R.
\newblock Leveraging constraint logic programming for neural guided program
  synthesis.
\newblock 2018.

\bibitem[Zhou \& Li(2005)Zhou and Li]{zhou2005tri}
Zhou, Z.-H. and Li, M.
\newblock Tri-training: Exploiting unlabeled data using three classifiers.
\newblock \emph{IEEE Transactions on Knowledge \& Data Engineering}, \penalty0
  (11):\penalty0 1529--1541, 2005.

\end{thebibliography}

\newpage
\newpage

\appendix

\newpage
\clearpage

\section{Code Availability}

All code is available at \url{https://github.com/yangkevin2/icml2020-stochastic-iterative-target-augmentation}.

\section{Toy Model}\label{toy}

We investigate the performance of our model in a toy setting, as follows. The discrete output space $\Omega$ is the set of points $(x_1, x_2)$ such that $-1 \leq x_i \leq 3$ and $x_i$ is a multiple of 0.05 for each $i=1,2$. We assume that there is no input, that is, we are operating in the unconditional setting. We define our constraint set to be the unit ball, so a new sample will pass our filter if and only if is in the unit ball. Our (nonparametric) prior is estimated using $k$-nearest neighbors density estimation on the dataset $\mathcal{D}$ with $k=50$, where $\mathcal{D}$ is initialized to be the set of points in $\Omega$ where both coordinates are multiples of 0.5 (in order to achieve a more even prior distribution over $\Omega$, even though some of these initial points are outside the constraint set). 

\begin{figure}
    \centering
    \includegraphics[width=0.475\textwidth]{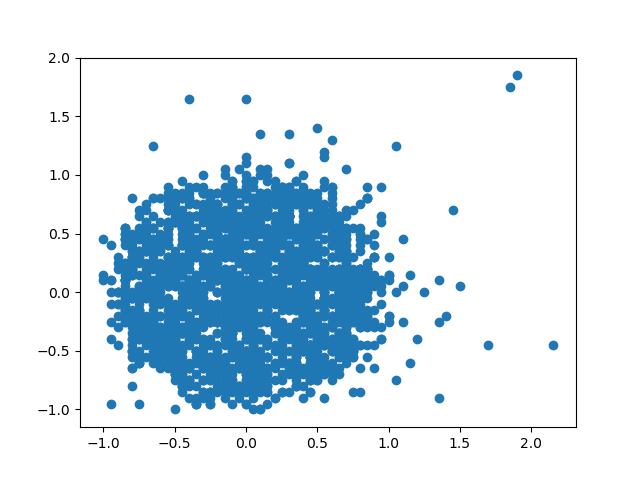}
    \caption{Distribution of samples at the end of toy model training. The unit ball is the desired constraint set, while the discrete space of possible samples is the points in $[-1, 3] \times [-1, 3]$ where both coordinates are multiples of 0.05.}
    \label{fig:toy_plot}
\end{figure}

We draw samples using Metropolis-Hastings~\citep{hastings1970monte} with interval of 50 steps between samples, using a uniform distribution over $\Omega$ as the proposal distribution. Upon drawing a sample, we add it to $\mathcal{D}$ if it lies in the unit ball. We repeat for a total of 20000 samples; adding the correct samples to $\mathcal{D}$ corresponds to our iterative augmentation and training procedure. Finally, for evaluation, we sample an additional 2000 samples (without filtering) and plot them in Figure \ref{fig:toy_plot}. Nearly all of the samples lie in the desired constraint set. 

\FloatBarrier

\section{Further Theoretical Analysis in Simplified Setting}\label{algo_behavior}

We analyze our method further in a simplified setting in order to understand our method's ability to produce diverse outputs. In particular, compared to Section \ref{theory} in the main text, we will now drop the input $X$, effectively switching to the unconditional setting. The constraint $\vc$ then depends only on $Y$. While producing diverse outputs is most important in the unconditional generation setting, our analysis here applies to the conditional setting as well, as we can view the conditional setting as a separate unconditional generation problem for each individual input. 





We will demonstrate that our method indeed yields a diverse distribution over correct outputs in a simplified nonparametric, non-stochastic setting. In this setting, our model $P$ has unlimited capacity, simulating an arbitrarily complex neural model in practice. Let $\mathcal{A} = \{Y: \vc = 1 | Y\}, \mathcal{B} = \{Y: \vc = 0 | Y\}$. Starting with a base distribution $P^{(0)}$, our objective will be to iteratively maximize $\log P(\mathcal{A})$, the log-probability that a sample from $P$ lies in $\mathcal{A}$. We also add a KL-divergence penalty to keep $P^{(t+1)}$ close to $P^{(t)}$ because in practice, we make only a limited update to our model distribution in each training iteration, dependent on learning rate. Thus, fixing some $\lambda > 0$, we update $P$ according to:

\begin{equation}\label{uc_update}
    P^{(t+1)} = \argmax_P \log P(\mathcal{A}) - \lambda KL(P||P^{(t)})
\end{equation}

where the argmax is over all possible models (distributions) $P$. We characterize $P^{(t+1)}$ by the following proposition, whose proof we defer to Appendix \ref{proofs}:

\begin{proposition}\label{npns_conv}
Assume $P^{(0)}$ has nonzero support on $\mathcal{A}$ and $\mathcal{B}$. Let $P^{(t)}(Y)$ be the probability of sampling molecule $Y$ from $P^{(t)}$, and $P^{(t)}(\mathcal{A})$ the probability that a given sample lies in $\mathcal{A}$. For any $\lambda > 0$, when updating $P$ according to Equation \ref{uc_update}, we have for all timesteps $t$ and molecules $Y$:
\begin{equation}\label{update_form}
    P^{(t)}(Y) = \alpha^{(t)} \frac{P^{(0)}(Y)}{P^{(0)}(\mathcal{A})} \mathbf{1}_{Y \in \mathcal{A}} + (1-\alpha^{(t)}) \frac{P^{(0)}(Y)}{P^{(0)}(\mathcal{B})} \mathbf{1}_{Y \in \mathcal{B}}
\end{equation}
for some sequence $\{\alpha^{(t)}\} \in [0,1]$. Moreover, the sequence $\{\alpha^{(t)}\}$ converges to 1, with $\alpha^{(t)} \geq 1 - \epsilon$ for $\epsilon > 0$ whenever $t \geq - \lambda \log(\epsilon\alpha^{(0)})$.
\end{proposition}

From Proposition \ref{npns_conv}, we observe that the converged model $P^{(\infty)}$ assigns probability to each output proportional to $P^{(0)} p(\vc=1 | Y)$. We conclude that in this simplified setting, if our starting distribution $P^{(0)}$ is reasonably diverse (for example, a randomly initialized neural generator), the resulting converged $P^{(\infty)}$ will be a diverse distribution over $\mathcal{A}$. 

\textbf{Remark.} In practice, since molecular structures are discrete and the distribution may be peaked, it is important to properly deal with repeated samples during our augmentation step. Thus we sample targets proportional to $P^{(t)} p(\vc = 1 |Y)$ \textit{without replacement}. This diverges from our theory, which corresponds to sampling \textit{with replacement}: the KL-divergence penalty encourages $P^{(t+1)}$ to assign probability proportional to $P^{(t)}$, rather than uniform probability across $\mathcal{A}$. In the limit as the number of samples goes to infinity, sampling targets without replacement is preferred: this encourages $P^{(\infty)}$ to be uniform over the set $\mathcal{A}$.

Lastly, we note that our analysis here applies in principle to the conditional setting as well, viewing each input precursor as a separate unconditional design task.

\section{Proof of Proposition \ref{npns_conv}}\label{proofs}

We now prove Proposition \ref{npns_conv}, reproduced below for convenience. 

\textbf{Proposition 1} \textit{(a) Assume $P^{(0)}$ has nonzero support on $\mathcal{A}$ and $\mathcal{B}$. Let $P^{(t)}(Y)$ be the probability of sampling molecule $Y$ from $P^{(t)}$, and $P^{(t)}(\mathcal{A})$ the probability that a given sample lies in $\mathcal{A}$. For any $\lambda > 0$, when updating $P$ according to Equation \ref{uc_update}, we have Equation \ref{update_form} for all timesteps $t$ and molecules $Y$:}
\begin{equation*}
    P^{(t)}(Y) = \alpha^{(t)} \frac{P^{(0)}(Y)}{P^{(0)}(\mathcal{A})} \mathbf{1}_{Y \in \mathcal{A}} + (1-\alpha^{(t)}) \frac{P^{(0)}(Y)}{P^{(0)}(\mathcal{B})} \mathbf{1}_{Y \in \mathcal{B}}
\end{equation*}
\textit{for some sequence $\{\alpha^{(t)}\} \in [0,1]$}

\textit{(b) Moreover, the sequence $\{\alpha^{(t)}\}$ converges to 1, with $\alpha^{(t)} \geq 1 - \epsilon$ for $\epsilon > 0$ whenever $t \geq - \lambda \log(\epsilon\alpha^{(0)})$.}

\textbf{Proof (a)} Recall Equation \ref{uc_update}:

\begin{equation*}
    P^{(t+1)} = \argmax_P \log P(\mathcal{A}) - \lambda KL(P||P^{(t)})
\end{equation*}

We first prove that the optimal $P$ exists and takes the stated form. Note that it suffices to prove the statement with $P^{(0)}(Y)$ replaced by $P^{(t)}(Y)$, as we can use induction. Each timestep $t$ simply results in a reweighting of the sets $\mathcal{A}$ and $\mathcal{B}$ by updating $\alpha$. 

Define $h(P) = \log P(\mathcal{A}) - \lambda KL(P||P^{(t)})$, and define a $P$ of the form given in Equation \ref{update_form} as \textit{proportionality-preserving}, or prop-preserving. First, we use a smoothing argument to show that for any non-prop-preserving $P_0$, there exists a prop-preserving $P^*$ such that $h(P_0) < h(P^*)$. 

By definition, 

\begin{align}
D(P) &\overset{def}{=} KL(P || P^{(t)}) \\
&= \sum_Y P(Y) \log P(Y) - P(Y) \log P^{(t)}(Y)
\end{align}

Taking the derivative with respect to $P(Y_0)$ for fixed $Y_0$:

\begin{align}
    \frac{dD(P)}{dP(Y_0)} &= 1 + \log P(Y_0) - \log P^{(t)}(Y_0)\\
    &= 1 + \log \frac{P(Y_0)}{P^{(t)}(Y_0)}\label{q_deriv}
\end{align}

Now for any $P_0$, let $\alpha_0$ be the weight it assigns to $\mathcal{A}$, and let $P^*_{\alpha_0}$ be the prop-preserving $P^*$ with parameter $\alpha_0$. For all $Y_0 \in \mathcal{A}$, because $P^*_{\alpha_0}$ is prop-preserving, $ \frac{P^*(Y_0)}{P^{(t)}(Y_0)}$ is equal to some constant $c$. Hence, $\frac{dD(P^*_{\alpha_0})}{dP^*_{\alpha_0}(Y_0)}$ is a constant $k$ for all $Y_0 \in \mathcal{A}$. 

Consider next the sets $\mathcal{A}_s$ and $\mathcal{A}_b$ which are the subsets of $\mathcal{A}$ where $\frac{P_0(Y_0)}{P^{(t)}(Y_0)} < c$ and $\frac{P_0(Y_0)}{P^{(t)}(Y_0)} > c$, respectively. Since $P_0$ and $P^*_{\alpha_0}$ assign the same probability to $\mathcal{A}$ as a whole, we have:

\begin{equation}
    P_0(\mathcal{A}_s) + P_0(\mathcal{A}_b) = P^*_{\alpha_0}(\mathcal{A}_s) + P^*_{\alpha_0}(\mathcal{A}_b) \label{eqmass}
\end{equation}

However, as the $\log$ function is strictly increasing, from \ref{q_deriv} we see that $\frac{dD(P_0)}{dP_0(Y_0)} < k$ whenever $\frac{P_0(Y_0)}{P^{(t)}(Y_0)} < c$ (i.e. $Y_0 \in \mathcal{A}_s$) and vice versa when $\frac{dD(P_0)}{dP_0(Y_0)} > k$ (i.e. $Y_0 \in \mathcal{A}_b$). Hence for $Y_0 \in \mathcal{A}_s$, by the mean value theorem we have that replacing $P_0(Y_0)$ with $P^*_{\alpha_0}(Y_0)$ would increase $D(P_0)$ by less than $k (P^*_{\alpha_0}(Y_0) - P_0(Y_0))$. Doing this replacement for all $Y_0 \in \mathcal{A}_s$ thus increases $D(P_0)$ by less than $k (P^*_{\alpha_0}(\mathcal{A}_s) - P_0(\mathcal{A}_s))$. Similarly, replacing $P_0(Y_0)$ with $P^*_{\alpha_0}(Y_0)$ for all $Y_0 \in \mathcal{A}_b$ decreases $D(P_0)$ by more than $k (P_0(\mathcal{A}_b) - P^*_{\alpha_0}(\mathcal{A}_b))$. 

However, from rearranging Equation \ref{eqmass} we have that $P_0(\mathcal{A}_s) - P^*_{\alpha_0}(\mathcal{A}_s) = - (P_0(\mathcal{A}_b) -  P^*_{\alpha_0}(\mathcal{A}_b))$. Therefore, replacing all values of $P_0(Y_0)$ with $P^*_{\alpha_0}(Y_0)$ for $Y_0 \in \mathcal{A}$ cannot increase the value of $D(P_0)$. Moreover, if $\mathcal{A}_s$ and $\mathcal{A}_b$ were nonempty, then $D(P_0)$ strictly decreases. 

We can repeat the same argument as above for the probability mass in $\mathcal{B}$. If $P_0$ is not prop-preserving, then either $\mathcal{A}_s$ and $\mathcal{A}_b$ are nonempty or the corresponding sets for $\mathcal{B}$ are nonempty. We conclude that for any non-prop-preserving $P_0$ there exists a prop-preserving $P^*_{\alpha_0}$ achieving a strictly lower value of $D(P) = KL(P || P^{(t)})$. Since $h$ places negative weight on $D(P)$, and our value replacements did not affect the value of $\log P_0(\mathcal{A})$, we conclude that $P^*_{\alpha_0}$ achieves a strictly higher value of $h$ than does $P_0$. 
  
Next, observe that the space of possible prop-preserving $P^*$ is one-dimensional, parameterized by $\alpha \in [0, 1]$. Thus, we can define a function $h'(\alpha)$ as $h(P^*(\alpha))$. Both $\log P(\mathcal{A})$ and $-\lambda KL(P || P^{(t)}$ are upper-bounded by 0, so $h' \rightarrow -\infty$ as $\alpha \rightarrow 0$. If $P^{(t)}(\mathcal{B}) = 0$ then we have the maximum at $\alpha = 1$, otherwise $h' \rightarrow -\infty$ as $\alpha \rightarrow 1$ as well. Since $h'$ is continuous and in fact strictly concave in $\alpha$ (due to strict concavity of $\log$ and convexity of $KL$), we conclude that $h'(\alpha)$ attains its unique maximum for some $\alpha^* \in [0, 1]$. Then since we showed previously that every non-prop-presering $P_0$ achieves a value of $h(P_0)$ at strictly less than that of some prop-preserving $P^*$, we conclude that a unique $P^*$ maximizing $h$ indeed exists and is prop-preserving. $\square$ 

\textbf{Proof (b)} We now show that the sequence $\{\alpha^{(t)}\}$ converges to 1. Since we assumed $P^{(0)}$ has nonzero support on $\mathcal{A}$, we know that $\alpha^{(0)} > 0$. If $P^{(0)}(\mathcal{B}) = 0$, then we are trivially done. So henceforth we can assume $\alpha^{(0)} \in (0, 1)$.

Noting that $\log P^*_{\alpha^{(t+1)}}(\mathcal{A}) = \alpha^{(t+1)}$ and $\log P^*_{\alpha^{(t+1)}}(\mathcal{B}) = 1- \alpha^{(t+1)}$, we have:

\begin{align}
    &KL(P^*_{\alpha^{(t+1)}} || P^{(t)}) \nonumber\\
    &= KL(P^*_{\alpha^{(t+1)}} || P^*_{\alpha^{(t)}})\\
    &= \sum_{Y_0 \in \mathcal{A}} P^*_{\alpha^{(t+1)}}(Y_0) \log \frac{P^*_{\alpha^{(t+1)}}(Y_0)}{P^*_{\alpha^{(t)}}(Y_0)}\\
    &\quad+ \sum_{Y_0 \in \mathcal{B}} P^*_{\alpha^{(t+1)}}(Y_0) \log \frac{P^*_{\alpha^{(t+1)}}(Y_0)}{P^*_{\alpha^{(t)}}(Y_0)}\nonumber\\
    &= P^*_{\alpha^{(t+1)}}(\mathcal{A}) \log \frac{\alpha^{(t+1)}}{\alpha^{(t)}} + P^*_{\alpha^{(t+1)}}(\mathcal{B}) \log \frac{1-\alpha^{(t+1)}}{1-\alpha^{(t)}}\\
    &= \alpha^{(t+1)}\log \frac{\alpha^{(t+1)}}{\alpha^{(t)}} + (1-\alpha^{(t+1)})\log \frac{1-\alpha^{(t+1)}}{1-\alpha^{(t)}}
\end{align}

We are now ready to take the derivative of $h$ with respect to $\alpha^{(t+1)}$:

\begin{align}
    &\frac{dh(P^*_{\alpha^{(t+1)}})}{d\alpha^{(t+1)}} \\
    &= \frac{1}{\alpha^{(t+1)}} - \lambda (\log \alpha^{(t+1)} + 1 - \log \alpha^{(t)}) \\
    &\quad-\lambda (\log (1-\alpha^{(t)}) - \log (1-\alpha^{(t+1)}) - 1)\nonumber\\
    &= \frac{1}{\alpha^{(t+1)}} - \lambda \log \frac{\alpha^{(t+1)}}{1-\alpha^{(t+1)}} + \lambda\log \frac{\alpha^{(t)}}{1-\alpha^{(t)}}
\end{align}

Observe that $\alpha$ is trivially nondecreasing: if $\alpha^{(t+1)} < \alpha^{(t)}$, then $\log P(\mathcal{A})$ decreases while $KL(P||P^{(t)})$ increases when comparing $P^*_{\alpha^{(t+1)}}$ and $P^*_{\alpha^{(t)}}$, as $P^{(t)} = P^*_{\alpha^{(t)}}$.

Moreover, the derivative $\frac{dh(P^*_{\alpha^{(t+1)}})}{d\alpha^{(t+1)}}$ is positive at $\alpha^{(t+1)} = \alpha^{(t)}$, so in fact $\alpha$ is strictly increasing. Since $h$ is continuous, we have either $\alpha^{(t+1)} = 1$ or $\frac{dh(P^*_{\alpha^{(t+1)}})}{d\alpha^{(t+1)}} = 0$. Solving the latter equation gives us 

\begin{align}
    \lambda\log \frac{\alpha^{(t)}}{1-\alpha^{(t)}} + \frac{1}{\alpha^{(t+1)}}&= \lambda\log \frac{\alpha^{(t+1)}}{1-\alpha^{(t+1)}}\\
    \frac{\alpha^{(t)}}{1-\alpha^{(t)}} e^{\frac{1}{\lambda\alpha^{(t+1)}}} &= \frac{\alpha^{(t+1)}}{1-\alpha^{(t+1)}} \label{before_ineq}
\end{align}

Now suppose for the sake of contradiction that $\{\alpha^{(t)}\}$ does not converge to 1, i.e. for some fixed $C < 1$, $\alpha^{t} < C$ for all $t$ . Then from \ref{before_ineq} we see that

\begin{align}
    \frac{\alpha^{(t)}}{1-\alpha^{(t)}} e^{\frac{1}{\lambda C}} &\leq \frac{\alpha^{(t+1)}}{1-\alpha^{(t+1)}}\label{bigger}
\end{align}


Finally, since $\frac{1}{\lambda C} > 0$, we have 

\begin{align}
    e^{\frac{1}{\lambda C}} > 1 \label{bound_power}
\end{align}

We conclude from \ref{bigger} and \ref{bound_power} that $\frac{\alpha^{(t)}}{1-\alpha^{(t)}}$ is exponentially increasing over time. This contradicts that $\alpha^{t} < C < 1$ for some fixed $C$ for all $t$; therefore, the sequence $\{ \alpha^{(t)}\}$ must converge to 1.

Finally, we analyze the rate of convergence. Suppose we want $\alpha^{(t)} \geq 1 - \epsilon$ for some $\epsilon > 0$, i.e., $\frac{\alpha^{(t)}}{1 - \alpha^{(t)}} \geq \frac{1 - \epsilon}{\epsilon}$. From \ref{bigger} we see when $\alpha < C$, $\frac{\alpha}{1-\alpha}$ is exponentially growing by a factor of at least $e^{\frac{1}{\lambda C}}$ with each timestep. Here, we have $C = 1-\epsilon$. Therefore, we have:

\begin{align}
    \frac{\alpha^{(t)}}{1 - \alpha^{(t)}} \geq \left( \frac{\alpha^{(0)}}{1 - \alpha^{(0)}} \right)e^{\frac{t}{\lambda (1-\epsilon)}}
\end{align}

From this, we see that to achieve $\frac{\alpha^{(t)}}{1 - \alpha^{(t)}} \geq \frac{1 - \epsilon}{\epsilon}$, it suffices to have:

\begin{align}
    \left( \frac{\alpha^{(0)}}{1 - \alpha^{(0)}} \right)e^{\frac{t}{\lambda (1-\epsilon)}} \geq \frac{1 - \epsilon}{\epsilon}
\end{align}

Rearranging gives us the following sufficient condition for $\alpha^{(t)} \geq 1 - \epsilon$:

\begin{align}
    t \geq \lambda(1-\epsilon) \log\left( \frac{(1-\epsilon)(1-\alpha^{(0)})}{\epsilon\alpha^{(0)}} \right)
\end{align}

Loosening the condition by observing $1-\epsilon < 1$ and $1-\alpha^{(0)} < 1$ gives us our desired $t \geq - \lambda \log(\epsilon\alpha^{(0)})$, although of course this bound is not tight. $\square$

\begin{figure*}[hbt!]
    \centering
    \begin{subfigure}[t]{0.475\textwidth}
        \centering
        \includegraphics[height=1.6in]{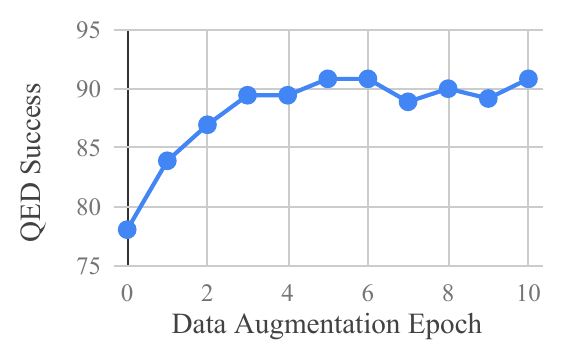}

    \end{subfigure}%
    \hfill
    \begin{subfigure}[t]{0.475\textwidth}
        \centering
        \includegraphics[height=1.6in]{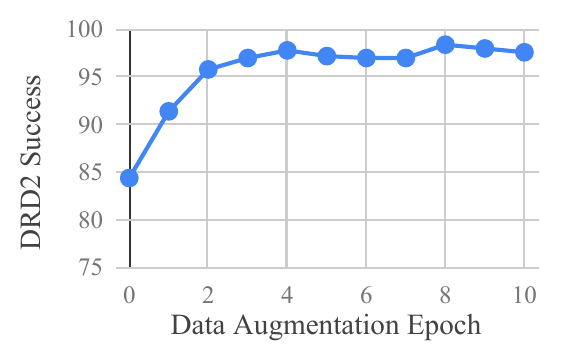}
    \end{subfigure}
    \caption{\textbf{Left}: Success rate for VSeq2Seq+ on validation set for each epoch of iterative target augmentation on conditional QED task. \textbf{Right}: Same plot for DRD2. For each plot, the far left point indicates the performance of the bootstrapped model.}
    \label{fig:learning_curves}
\end{figure*}

\section{Additional Experimental Details} 

\subsection{Implementation and Hyperparameters} \label{hyperparams}
Our augmented models share the same hyperparameters as their baseline counterparts in all cases.

In the molecular design conditional case, for VSeq2Seq we use batch size 64, embedding and hidden dimension 300, VAE latent dimension 30, and an LSTM with depth 1 (bidirectional in the encoder, unidirectional in the decoder). For models using stochastic iterative target augmentation, $n_1$ is set to $5$ and $n_2$ is set to $10$, while for the baseline models we train for $20$ epochs  (corresponding to $n_1 = 20, n_2=0$). The HierGNN model shares the same hyperparameters as in \citet{jin2019multi}.

In the unconditional setting, our VSeq model uses the same hyperparameters as the conditional-case VSeq2Seq model, while for the REINVENT baseline we use \citet{olivecrona2017molecular}'s default settings. Both models have approximately 4 million trainable parameters to facilitate fair comparison. We set $n_1$ to $1$ and $n_2$ to 50, and train the VSeq baseline model for $50$ epochs. We also discard the gold data altogether after the initial bootstrapping phase, as we find that this improves model performance. For the REINVENT baseline, we train their prior model for the recommended number of steps, and then finetune using their RL method until convergence. We additionally searched over their $\sigma$ hyperparameter, although we found that this did not significantly affect performance on either the QED or DRD2 tasks, so our final runs use the default value of 20. 

For the training time and prediction time filtering parameters, we set $K=4$, $C=200$, and $L=10$ for both the QED and DRD2 tasks, in both the conditional and unconditional cases. Although we ran experiments with different values of $K$, we found that the change did not significantly affect performance unless $K$ was too small; see Appendix \ref{more_molopt_exp}.

For the Karel program synthesis task, we use the same hyperparameters as the MLE baseline model in \citet{bunel2018leveraging}. Our augmented model shares the same hyperparameters. We use a beam size of $64$ at test time, the same as the MLE baseline, but simply sample programs from the decoder distribution when running iterative target augmentation during training. The baseline model is trained for $100$ epochs, while for the model employing iterative target augmentation we train as normal for $n_1 = 15$ epochs followed by $n_2 = 50$ epochs of iterative target augmentation. Due to the large size of the full training dataset, in each epoch of iterative augmentation we use $\frac{1}{10}$ of the dataset, so in total we make $5$ passes over the entire dataset. 

For the training time and prediction time filtering parameters, we set $K=4$, $C=50$, and $L=10$.

All code is in PyTorch~\citep{paszke2017automatic}.

\subsection{Dataset Sizes}

In Table \ref{tab:data_size} we provide the training, validation, and test set sizes for all of our tasks. Note that the validation and test sizes are relevant only for the conditional case. For each task we use the same splits as our baselines. 

\begin{table}[hbt!]
\centering
\begin{tabular}{@{}lrrr@{}}
\toprule
\textbf{Task}  & \textbf{Training Set} & \textbf{Validation Set} & \textbf{Test Set} \\ \midrule
QED   & 88306        & 360            & 800      \\ 
DRD2  & 34404        & 500            & 1000     \\ 
Karel & 1116854 & 2500 & 2500 \\ \bottomrule
\end{tabular}
\caption{Number of examples in training, validation, and test sets for each task.}
\label{tab:data_size}
\end{table}

The QED data is obtained from filtering ZINC~\citep{sterling2015zinc}, while the DRD2 data is obtained from ZINC and \citet{olivecrona2017molecular}. Our complete datasets are included together with our code submission. 

\subsection{Learning Curves}

In Figure \ref{fig:learning_curves}, we provide the validation set performance per augmentation epoch for our VSeq2Seq+ model on both the QED and DRD2 conditional tasks.

\begin{figure*}[t]
    \centering
    \begin{subfigure}[b]{0.475\textwidth}
        \centering
        \includegraphics[width=\textwidth]{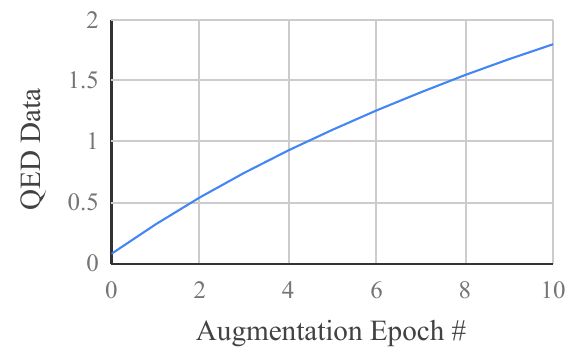}
        \caption{QED Conditional}
        \label{fig:unique_data_over_timea}
    \end{subfigure}
    \hfill
    \begin{subfigure}[b]{0.475\textwidth}  
        \centering 
        \includegraphics[width=\textwidth]{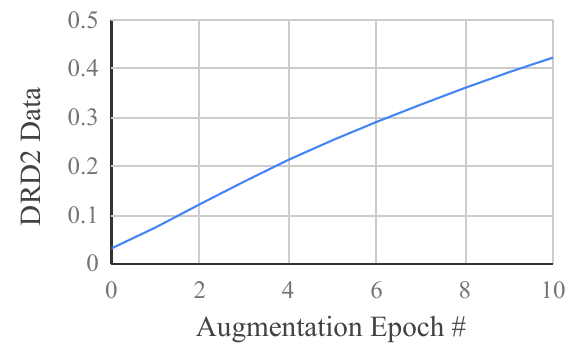}
        \caption{DRD2 Conditional}
        \label{fig:unique_data_over_timeb}
    \end{subfigure}
    \vskip\baselineskip
    \begin{subfigure}[b]{0.475\textwidth}   
        \centering 
        \includegraphics[width=\textwidth]{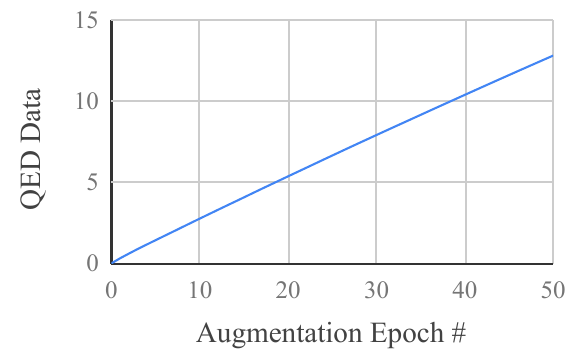}
        \caption{QED Unconditional}
        \label{fig:unique_data_over_timec}
    \end{subfigure}
    \hfill
    \begin{subfigure}[b]{0.475\textwidth}   
        \centering 
        \includegraphics[width=\textwidth]{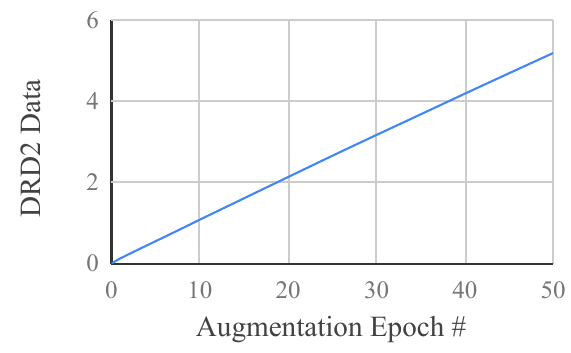}
        \caption{DRD2 Unconditional}
        \label{fig:unique_data_over_timed}
    \end{subfigure}
    \caption{Graphs of the cumulative number of unique training pairs our augmented sequence-based model has seen by the time of each augmentation epoch, on both QED and DRD2 tasks in both conditional and unconditional settings. All vertical axis scales in millions.}
    \label{fig:unique_data_over_time}
\end{figure*}

\FloatBarrier

\subsection{Unique Data Seen Over Time}

In Figure \ref{fig:unique_data_over_time}, we show the cumulative number of unique data points seen during augmentation epochs. The four subplots show the QED and DRD2 tasks for both the VSeq2Seq+ model in the conditional setting as well as the VSeq+ model in the unconditional setting. Even after several epochs, the number of unique data points is still increasing in all cases. Due to the large number of additional data points, we find that in both settings, empirical model performance at test time is limited more by the discrepancy between the proxy predictor and the ground truth evaluator than by the number of new data points seen. This is evidenced by the near-perfect performance we observe for both VSeq2Seq+ and VSeq+ when evaluated using the proxy predictor. 

\begin{table}[]
\centering
\begin{tabular}{@{}lll@{}}
\toprule
\multicolumn{3}{c}{\textit{\textbf{Conditional}}}   \\
\toprule
\textbf{Model}         & \textbf{QED}     & \textbf{DRD2}    \\ \midrule
VSeq2Seq      & 1.34    & \phantom{2}7.74    \\
VSeq2Seq+     & 1.28    & \phantom{2}7.10    \\
\bottomrule
\multicolumn{3}{c}{\textit{\textbf{Unconditional}}} \\ \toprule
\textbf{Model}         & \textbf{QED}     & \textbf{DRD2}    \\ \midrule
VSeq          & 3.21    & 12.45   \\
REINVENT      & 4.79    & 19.81   \\
VSeq+         & 3.33    & 10.86 \\ \bottomrule
\end{tabular}
\caption{FCD evaluation of baselines and augmentations on two datasets in both conditional and unconditional settings; in isolation, lower is better. Our augmentation method maintains similar FCD between outputs and gold targets compared to the baseline on QED, and decreases the distance on DRD2, while substantially improving the success rate and diversity of modifications. By contrast, the reinforcement-learning based REINVENT method greatly increases the FCD on both QED and DRD2 in the unconditional setting. }
\label{tab:fcd}
\end{table}

\begin{table*}[t]
\centering
\begin{tabular}{@{}lcccc@{}}
\toprule
\textbf{Model} & \textbf{QED Succ.}    & \textbf{QED Div.} & \textbf{DRD2 Succ.} & \textbf{DRD2 Div.}       \\ \midrule
\textit{VSeq2Seq+, K=2} & 85.1          & 0.453        & 95.9           & 0.327 \\ 
\textit{VSeq2Seq+, K=4} & 89.0            & 0.470         & 97.2           & 0.361 \\ 
\textit{VSeq2Seq+, K=8} & 88.4          & 0.480         & 97.6           & 0.373 \\ \bottomrule
\end{tabular}
\caption{Performance of our model VSeq2Seq+ in the conditional setting with different values of $K$. All other experiments use $K = 4$.}
\label{tab:effect_k}
\end{table*}

\begin{table*}[t]
\centering
\begin{tabular}{@{}lcccc@{}}
\toprule
\textbf{Model} & \textbf{QED Succ.}             & \textbf{QED Div.} & \textbf{DRD2 Succ.} & \textbf{DRD2 Div.}       \\ \midrule
\textit{VSeq2Seq+ }              & 89.0            & 0.470 & 97.2           & 0.361 \\ 
\textit{VSeq2Seq+, keep-targets} & 89.8          & 0.465        & 97.6           & 0.363 \\ \bottomrule
\end{tabular}
\caption{Performance in conditional setting of our proposed augmentation scheme, VSeq2Seq+, compared to an alternative version (VSeq2Seq+, keep-targets) which keeps all generated targets and continually grows the training dataset.}
\label{tab:keep_translations}
\end{table*}


\subsection{Frechet ChemNet Distance Analysis}\label{fcd}

As another evaluation of our model on a metric it was not optimized for, we further evaluate Frechet ChemNet Distance (FCD)~\citep{preuer2018frechet} between model outputs and a reference set of gold targets for both the QED and DRD2 tasks, in both the conditional and unconditional settings. FCD is the molecular analogue of Frechet Inception Distance for images~\citep{heusel2017gans}, measuring distributional distance. Considering the FCD metric in isolation, we prefer models whose outputs have lower distributional distance with the reference set. 

In Table \ref{tab:fcd}, we observe that on the QED task our model and the baseline perform similarly. On DRD2, our augmentation method is quite successful at decreasing the distributional distance, where the distributional distances between the training targets and the reference gold targets leave more room for improvement compared to QED. Thus our method improves over the baseline in success and diversity (our main metrics in the paper) while also performing equal or better by FCD. By contrast, we observe that the REINVENT baseline heavily degrades performance on the FCD metric compared to the baseline on both tasks. 

\subsection{Further Molecular Design Experiments}\label{more_molopt_exp}

In the conditional case, we experiment with the effect of modifying $K$, the number of new targets added per precursor during each training epoch. In all other experiments we have used $K=4$. Since taking $K=0$ corresponds to the base non-augmented model, it is unsurprising that performance may suffer when $K$ is too small. However, as shown in Table \ref{tab:effect_k}, at least in the conditional case there is relatively little change in performance for $K$ much larger than $4$.

We also experiment with a version of our method which continually grows the training dataset by keeping all augmented targets, instead of discarding new targets at the end of each epoch. We chose the latter version for our main experiments due to its closer alignment to our EM motivation. However, we demonstrate in Table \ref{tab:keep_translations} that performance gains from continually growing the dataset are small to insignificant in our conditional molecular design tasks. 

\FloatBarrier

\begin{table*}[t]
\centering
\begin{tabular}{@{}lcccccc@{}}
\toprule
\textbf{Model} & \textbf{Train+} & \textbf{Test+} & \textbf{QED Succ.}  & \textbf{QED Uniq.}& \textbf{DRD2 Succ.} &\textbf{DRD2 Uniq.}\\ \midrule
VSeq         & {\color{red}\xmark }               & {\color{red}\xmark }                        & 62.4     & 0.499         & 51.4    & 0.221          \\
\textit{VSeq(test)}  &{\color{red}\xmark }               & {\color{green}\cmark }                       & 96.5  & 0.732     &   92.4             &  0.338           \\
\textit{VSeq(train)} & {\color{green}\cmark }               & {\color{red}\xmark }                        & 95.3  & 0.953     & 92.5                  & 0.924          \\
\textit{VSeq+}        & {\color{green}\cmark }               & {\color{green}\cmark }                       & 95.8 & 0.957     & 92.8                  & 0.927         \\
\textit{VSeq(dupe)}        & {\color{green}\cmark }               & {\color{green}\cmark }                       & 93.2     & 0.886         & 83.9         & 0.511         \\\bottomrule
\end{tabular}
\caption{Ablation analysis of filtering at training and test time for unconditional molecular generation. ``Train+'' indicates a model whose training process uses data augmentation according to our framework. ``Test+'' indicates a model that uses the external filter at prediction time to discard candidate outputs which fail to pass the filter. }
\label{tab:uc_filter}
\end{table*}

\begin{figure*}[t!]
    \centering
    \begin{subfigure}[t]{0.475\textwidth}
        \centering
        \includegraphics[height=1.6in]{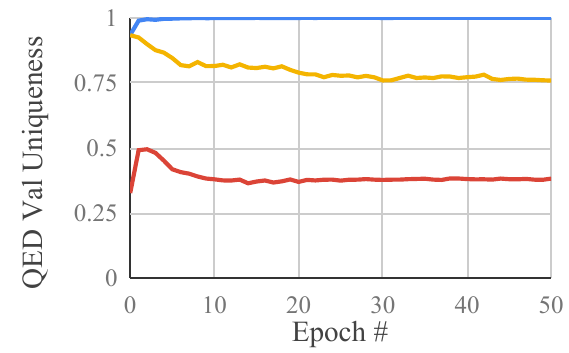}
    \end{subfigure}%
    \hfill
    \begin{subfigure}[t]{0.475\textwidth}
        \centering
        \includegraphics[height=1.6in]{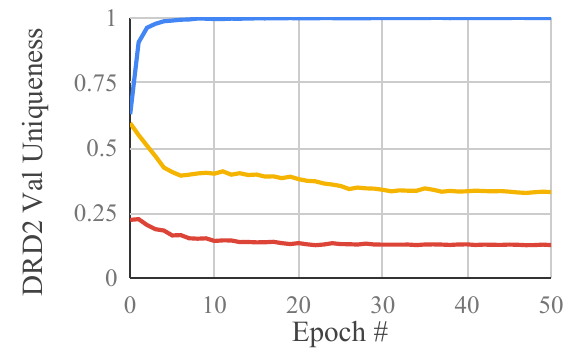}
    \end{subfigure}
    \vspace{-5pt}
    \caption{\textbf{Left}: Epoch number vs. uniqueness, evaluated with the Chemprop proxy predictor, for VSeq-based models on QED dataset. VSeq+, Vseq(dupe), and VSeq in blue, yellow, and red respectively. \textbf{Right}: Same plot for DRD2. Note that both VSeq(dupe) and VSeq+ are trained without iterative target augmentation for the initial epoch 0, and trained with augmentation thereafter.}
    \label{fig:full_uniqueness}
\end{figure*}

\subsection{Model Stability and Number of Runs}

We found that the reinforcement-learning based REINVENT model was sometimes unstable on our DRD2 dataset, resulting in wide variance in results between different runs. To confirm statistical significance, we ran VSeq+ and REINVENT 10 times each on this dataset, resulting in VSeq+ having higher uniqueness with p-value $0.003$ in a t-test.

All other models were highly stable and performed consistently between runs, particularly in the conditional setting. For our final experiments we ran all models 3 times in the unconditional setting, reporting mean metrics, and once in the conditional setting. 

\subsection{Unconditional Molecular Design Ablations}\label{uc_ablations}

In Table \ref{tab:uc_filter} we present an ablation analysis for the unconditional setting, similar to that for the conditional setting in the main text. We also analyze an ablation VSeq(dupe), an ablation of our stochastic iterative target augmentation method applied to VSeq. It samples targets with replacement during augmentation, unlike our full method which deduplicates. As suggested by our theoretical remark on the difference between sampling with and without replacement in Appendix \ref{algo_behavior}, VSeq(dupe) underperforms VSeq+. As Figure \ref{fig:full_uniqueness} demonstrates, its diversity eventually decreases over time.


Interestingly, VSeq(train) achieves nearly the same uniqueness score as VSeq+, indicating that the additional training targets from our stochastic iterative augmentation method are responsible for most if not all the gains over the baseline. In particular, even our ablation model VSeq(train) significantly outperforms the REINVENT baseline, demonstrating that our model's advantage over RL is not limited to our prediction-time filtering procedure. 

\subsection{Program Synthesis Ablations}

In Table \ref{tab:prog_filter} we provide the same ablation analysis that we provided in the main text for the conditional molecular design task, demonstrating that both training time iterative target augmentation as well as prediction time filtering are beneficial to model performance. We hypothesize that the effect of test-time filtering is relatively larger in program synthesis than in molecular design because checking correctness is easier in this domain. However, we note that even MLE(train), our model without prediction time filtering, outperforms the best RL method from \citet{bunel2018leveraging}.

\begin{table}[t]
\centering
\begin{tabular}{@{}lccc@{}}
\toprule
\textbf{Model}            & \textbf{Train+} & \textbf{Test+} & \textbf{Top-1 Generalization} \\ \midrule
MLE$^*$         & {\color{red}\xmark }               & {\color{red}\xmark }                        & 70.91                  \\
\textit{MLE(test)$^*$}  &{\color{red}\xmark }               & {\color{green}\cmark }                       & 79.61                    \\
\textit{MLE(train)} & {\color{green}\cmark }               & {\color{red}\xmark }                        & 77.92                  \\
\textit{MLE+}        & {\color{green}\cmark }               & {\color{green}\cmark }                       & \textbf{85.02}                   \\\bottomrule
\end{tabular}
\caption{Ablation analysis of filtering at training and test time for program synthesis. ``Train+'' indicates a model whose training process uses data augmentation according to our framework. ``Test+'' indicates a model that uses the external filter at prediction time. *Note that MLE and MLE(test) are based on an MLE checkpoint which underperforms the published result from \citet{bunel2018leveraging} by 1 point, due to training for fewer epochs.}
\label{tab:prog_filter}
\end{table}

\end{document}